\documentclass[journal]{IEEEtran}
\usepackage{amsmath,amsfonts}
\usepackage{algorithmic}
\usepackage{algorithm}
\usepackage{array}
\usepackage[caption=false,font=normalsize,labelfont=sf,textfont=sf]{subfig}
\usepackage{textcomp}
\usepackage{stfloats}
\usepackage{url}
\usepackage{verbatim}
\usepackage{graphicx}
\usepackage{multirow}
\usepackage{hyperref}

\renewcommand{\arraystretch}{1.25}

\usepackage{xcolor}
\definecolor{brightgreen}{rgb}{0.0, 0.8, 0.0}

\newcommand{\gcdots}{{\color{gray}\cdots}}
\newcommand{\gvdots}{{\color{gray}\vdots}}
\newcommand{\gddots}{{\color{gray}\ddots}}

\makeatletter
\def\bbordermatrix#1{\begingroup \m@th
  \@tempdima 4.75\p@
  \setbox\z@\vbox{%
    \def\cr{\crcr\noalign{\kern2\p@\global\let\cr\endline}}%
    \ialign{$##$\hfil\kern2\p@\kern\@tempdima&\thinspace\hfil$##$\hfil
      &&\quad\hfil$##$\hfil\crcr
      \omit\strut\hfil\crcr\noalign{\kern-\baselineskip}%
      #1\crcr\omit\strut\cr}}%
  \setbox\tw@\vbox{\unvcopy\z@\global\setbox\@ne\lastbox}%
  \setbox\tw@\hbox{\unhbox\@ne\unskip\global\setbox\@ne\lastbox}%
  \setbox\tw@\hbox{$\kern\wd\@ne\kern-\@tempdima\left[\kern-\wd\@ne
    \global\setbox\@ne\vbox{\box\@ne\kern2\p@}%
    \vcenter{\kern-\ht\@ne\unvbox\z@\kern-\baselineskip}\,\right]$}%
  \null\;\vbox{\kern\ht\@ne\box\tw@}\endgroup}
\makeatother

\usepackage{cuted}

\usepackage{tikz}
\usetikzlibrary{matrix}

\begin{document}

\title{Multi-stage warm started optimal motion planning for over-actuated mobile platforms}

\author{G.J. Paz-Delgado$^{1,3}$, C.J. Pérez-del-Pulgar$^{1}$, M. Azkarate$^{2}$, F. Kirchner$^{3}$, A. García-Cerezo$^{1}$
\thanks{This work has been partially funded by the European Space Agency, contract no. 4000118072/16/NL/LvH/gp and the European Comission, under the H2020 project CoRob-X with grant agreement 101004130.}
\thanks{$^{1}$G.J. Paz-Delgado, C.J. Pérez-del-Pulgar and A. García-Cerezo are with the Space Robotics Laboratory, Department of Systems Engineering and Automation, Universidad de Málaga, Andalucia Tech, 29070 Málaga, Spain
        {\tt\small  carlosperez@uma.es}}%
\thanks{$^{2}$M. Azkarate is with the Automation and Robotics Section, European Space Agency,
        Noordwijk, AZ 2201, The Netherlands
        {\tt\small martin.azkarate@esa.int}}
\thanks{$^{3}$G.J. Paz-Delgado and F. Kirchner are with the DFKI Robotics Innovation Center Bremen, 
        Robert-Hooke-Str. 1, 28359 Bremen, Germany
        {\tt\small frank.kirchner@dfki.de}}
        }

\markboth{}%
{Shell \MakeLowercase{\textit{et al.}}: A Sample Article Using IEEEtran.cls for IEEE Journals}


\maketitle

\begin{abstract}
This work presents a computationally lightweight motion planner for over-actuated platforms. 
For this purpose, a general state-space model for mobile platforms with several kinematic chains is defined, which considers non-linearities and constraints. 
The proposed motion planner is based on a sequential multi-stage approach that takes advantage of the warm start on each step. 
Firstly, a globally optimal and smooth 2D/3D trajectory is generated using the Fast Marching Method. 
This trajectory is fed as a warm start to a sequential linear quadratic regulator that is able to generate an optimal motion plan without constraints for all the platform actuators. 
Finally, a feasible motion plan is generated considering the constraints defined in the model. In this respect, the sequential linear quadratic regulator is employed again, taking the previously generated unconstrained motion plan as a warm start.
This novel approach has been deployed into the Exomars Testing Rover of the European Space Agency. This rover is an Ackermann-capable planetary exploration testbed that is equipped with a robotic arm.
Several experiments were carried out demonstrating that the proposed approach speeds up the computation time, increasing the success ratio for a martian sample retrieval mission, which can be considered as a representative use case of an over-actuated mobile platform.

\end{abstract}


\section{Introduction}

\IEEEPARstart{A}{utonomy} is a must-have feature for many mobile platforms nowadays, allowing to fulfill a given task with a minimal, and often unavailable human intervention. 
This enhancement is particularly key for mobile platforms, which could be defined as an actuated system without workspace limitations thanks to a non-fixed base, like terrestrial robots, Unmanned Aerial Vehicles (UAVs), satellites or planetary exploration vehicles.
These mobile platforms are sometimes teleoperated, which is helpful in changing scenarios. However, there is a growing interest in the research of novel methods to increase their autonomy level.

One of the key challenges is related to the motion planning problem.
It is a well-known problem in the literature, which could be defined as finding a feasible trajectory for each actuator of the platform to perform a goal task interacting with the real world \cite{Latombe1991RobotPlanning}. 
This entails many challenges.
For instance, Unmanned Ground Vehicles (UGVs) \cite{Alatise2020AMethods}, autonomous driving  \cite{Soni2018FormationSurvey} and planetary exploration \cite{Gerdes2020EfficientResources} vehicles usually have non-holonomic constraints on the mobile base movements. This means that, depending on the system state, the mobility is restricted to certain directions, commonly forwards and backwards.
Ground vehicles, Autonomous Surface Vehicles (ASVs) \cite{Wang2019Roboat:Waterways} and UAVs \cite{Kratky2021AutonomousVehicles} deal with non-static, dynamically changing scenarios. Hence, a requirement arises for the planner to consider multiple evolving time constraints, to avoid collisions with other movable elements in the scenario, i.e. people, other vehicles or animals.
Localization uncertainty in planetary exploration or submarine vehicles \cite{Huang2018EfficientVehicle} hinders the extraction of safe motions w.r.t the scenario.
Furthermore, the robustness of the planner for this applications has to be maximized, due to the difficulty of reparations or replacements once the mission is launched, such as happens with satellites \cite{Araguz2018ApplyingProspects}.
Besides, all of these systems have a common challenge for the motion planner: the computational effort. Mobile platforms computational resources are usually limited, even more for space or submarine qualified systems, or for small-sized ones like UAVs and sub-gram robots \cite{Pierre2019TowardRobots}.

On top of that, the  aforementioned challenges are even more difficult to be solved for complex systems, such as platforms that have a higher number of actuated joints than Cartesian Degrees of Freedom (DoF) to be controlled.
These are called over-actuated platforms, for instance, mobile manipulators, which are mobile platforms equipped with a robotic arm. 
The over-actuation entails the existence of infinite solutions for the motion planning problem, since the same goal can be reached differently in function of the selected joints to move, considering their physical properties, their limits or the characteristics of the scenario.
Nevertheless, this is actually an advantage if properly examined, since there are a great variety of solutions where to look for the one that best fits the problem needs. 

This search can be thoroughly performed by different methods, such as potential fields, artificial intelligence, optimization or probabilistic algorithms. As an example, a particular solution in the case of mobile manipulators is to decouple the mobile base and the manipulator plans. This divides the global over-actuated problem into two smaller non-redundant ones, moving the manipulator exclusively when the base is close to the goal, or when it could collide with the scenario \cite{Pilania2018MobileEnvironments}. 
However, this lacks of the efficiency that a coupled arm-base motion provides, thus, finer approaches generate full body, dynamic motion plans. 
For instance, non-holonomic constraints, manipulator singularities and mechanical limits are the main focus of the full body motion planner presented in \cite{Pajak2017Point-to-PointManipulators}, where a reactive algorithm modifies the motion of the robot when it is moving in the vicinity of obstacles.
Another typical example of a highly efficient motion planner dealing with over-actuation in complex static environments is the Optimized Hierarchical Mobile Manipulator Planner (OHMP), depicted in \cite{Li2020AManipulator}, which computes first a bidimensional path for the mobile base using Probabilistic RoadMaps (PRM) and then safe manipulator configurations using PRM again in a 2D projection of the 3D space.

Although tailored solutions like the aforementioned are suitable for particular platforms and situations, the most versatile motion planners for general, efficiency demanding use cases are the optimization based ones, which are characterized by searching for the best solution considering certain factors, like energy consumption, collision avoidance or time spent. 
In fact, optimization based motion planners are able to tackle smoothly the motion planning main challenges, particularly the over-actuation: since there are infinite combinations of joint movements to reach the goal, the optimization will find at least the locally best one considering the selected criteria. 
For instance, in \cite{Liao2019Optimization-basedRedundancy} an optimization twofold looped motion planner for mobile manipulators is proposed, with an outer loop to penalize inequality constraints, like joint limits or obstacles, and an inner loop to find the collision-free trajectory. 
Nonlinear system dynamics and constraints are considered within the Constrained Iterative Linear Quadratic Regulator (CILQR) \cite{Chen2019}, which solves the optimal control problem of planning the motion of an autonomously driving vehicle. 
To maximize the time available to perform scientific measurements on a spacecraft, in \cite{Preda2021OptimalProgramming} a sequential convex programming method generates optimal reorientation policies to maintain a comet within the field of view of the onboard instruments.
Looking for the fastest possible performance, in \cite{Foehn2021Time-optimalFlight} the time is included as a variable to be optimized, achieving an optimal motion plan for a quadrotor that significantly beats human pilots when benchmarked inside a static 3D scenario.
With the objective of tracking end effector poses or open doors, a Model Predictive Control (MPC) loop is integrated in \cite{Minniti2019} on a ball-balancing mobile manipulator, which is inherently unstable, by including a Sequential Linear Quadratic regulator (SLQ) into the MPC scheme. 
Platform non-holonomic constraints are included into the motion planner of a highly redundant mobile manipulator in \cite{Giftthaler2017} by using a constrained version of SLQ. 
Additionally, optimizing the gait switching times and the control inputs is also achieved by means of the Constrained SLQ in \cite{Farshidian2017}, applied to quadrupedal locomotion control and motion planning.

As can be observed, most of the aforementioned motion planners are integrated into a motion plan tracking scheme, which ensures the goal is correctly reached by recomputing the motion continuously online, thus compensating any external disturbances and errors not considered initially.
Multiple different schemes are found in the literature, being MPC the most useful and spread approach, capable of dealing with non-linearities, system dynamics, system redundancy (or over-actuation) and constraints on the states or inputs. 
Recent works have applied MPC to trajectory tracking of mobile robots \cite{Fnadi2021ConstrainedGrounds} and hexacopters \cite{Neunert2016}, mobile manipulation for transporting heavy objects \cite{Wang2021TransportingFramework}, with non-holonomic constraints \cite{Colombo2019ParameterizedApproach}, quadrupedal locomotion \cite{Farshidian2017} or even quadrupedal mobile manipulation \cite{Sleiman2021AManipulation}. 
Also known as receding horizon optimal control, MPC generally defines a short finite time horizon where the prediction is performed. This prediction time keeps receding, or moving forwards, until the goal is reached. 
Nevertheless, latest optimal controllers like Nonlinear Model Predictive Horizon (NMPH) \cite{Younes2021NonlinearGeneration}, based also on MPC, suggest that generating a complete reference trajectory for the full prediction horizon improves noticeably the smoothness of the solution, since the motion is optimized from the current state of the system until the terminal set point, not only for the next few time steps.
Additionally, the event-triggered Replanning presented in \cite{Luis2020OnlinePlanning} improves the smoothness of the robot movements in comparison to standard MPC approaches. This is achieved by only matching the reference for the motion planner with the platform current state when an external disturbance is detected. Hence, the jittering, generated by the actuation discontinuities, is removed.

All the optimization and tracking algorithms presented above have, nevertheless, a main drawback: they require high computational efforts, even more when considering the continuous replanning requirements inside the MPC schemes. 
This issue is caused mainly by the need of computing the system dynamics at each iteration, which consumes between 30\% and 90\% of the CPU time of many state-of-art motion planners \cite{Plancher2021AcceleratingFPGA}. Thus, it is critical for these algorithms to reduce as much as possible the average number of iterations until convergence, which is achievable by means of the so-called warm start. 
The objective of warm start is to provide the optimization algorithm with an initial solution of the problem, which is not necessarily feasible, but is used as a starting point that places the algorithm close to the convex area surrounding a local (or global) optimal solution. Hereby, the optimization is boosted to find the solution faster, in much fewer iterations. 
This is clearly demonstrated in \cite{Bitar2019Warm-StartedASVs} for trajectory planning of ASVs, where $A^*$ is used to warm start the optimal planner with the shortest path to reach the goal. This warm start method improves by 84\% the time spent in comparison to the same cold started optimal planner, when applied to under-actuated platforms. 
An improved version of $A^*$, $Hybrid\:A^*$ is used in \cite{Zhang2019AutonomousAvoidance} to warm start the trajectory of an Optimization-Based Collision Avoidance (OBCA) method. Applied to autonomous vehicle parking, again an under-actuated system, this warm start reduces by a factor of 4 the time of convergence in comparison to the same cold started OBCA.

Warm start also increases noticeably the success rate of iterative descent based motion planners, as shown in \cite{Lembono2020MemoryOptimization}, where different function approximation methods like \textit{k-Nearest Neighbor} (k-NN), \textit{Gaussian Process Regressor} (GPR) or \textit{Bayesian Gaussian Mixture Regression} (BGMR) are used altogether to warm start multiple trajectory optimizations in parallel. This increases greatly the number of success motion plans, reaching 71\% success rate when applied to the humanoid robot ATLAS with the goal of reaching a random Cartesian pose, although using a small quantity of time steps and without bringing the plan to the real platform.
Furthermore, a memory of motion is computed and stored using PRM offline in \cite{Mansard2018UsingController} to warm start the optimal controller during the motion, achieving a 93.5\% success rate of the solver in less than 5 iterations when applied to a planar simulated UAV, which means an improvement of approximately 5 times in the cost of the algorithm w.r.t. the cold started one, although it is a simple under-actuated system.
It is also remarkable the use of several layers to warm start the optimization, as shown in \cite{Ichnowski2020DeepPlanning}, where a first stage neural network estimates the trajectory horizon, used by a second stage neural network that estimates an initial trajectory, which is used to warm start the third stage: the optimization motion planner. Applied only to a fixed-base robotic arm, this multi-staged approach manages to reduce the computation time by 99\%. 

Seeing the foregoing contributions, it is clear that a proper warm start strongly accelerates the convergence of iterative descent optimal motion planners. 
Moreover, multi-staged warm start is arising as an interesting methodology to increase even to a greater extent the convergence speed. 
Nevertheless, the benefits of warm start on over-actuated systems are yet to be proven, which could improve noticeably the low convergence ratios and high computational cost when planning the motions of this type of systems, caused by their complexity, constraints and task requirements.
Thus, in this paper a general multi-staged warm started motion planner for over-actuated mobile platforms is proposed.
The first stage is a path planner that computes an initial trajectory for the mobile platform, the second stage takes that initial trajectory as warm start and solves only the unconstrained problem, and the solution of the unconstrained problem is used as the warm start of the third stage, which is a constrained optimization solver.

To validate the presented approach, firstly a generic state space model for over-actuated mobile platforms is proposed, including several kinematic chains, system dynamics and external disturbances.
Secondly, this general model is particularized to a double-Ackermann mobile manipulator, including specific constraints like the non-holonomic constraints of the mobile base or the actuator joint limits.
Additionally, a tailored replanning methodology is developed to ensure that the mobile manipulator properly tracks the planned motion, based on the event-triggered Replanning presented in \cite{Luis2020OnlinePlanning}.
Lastly, the results of a lab test campaign are shown, emulating a Sample Fetch Rover (SFR) sample retrieval scenario with ExoTeR (Exomars Testing Rover) in the Martian Analogue Testbed located at the Planetary Robotics Laboratory (PRL) of the European Space Agency (ESA).

The rest of the paper is organized as follows. In Section \ref{sec:algorithm} the motion planning algorithm is presented and explained in detail, including the generic space state model for over-actuated mobile platforms.
In Section \ref{sec:use_case} a particular use case of the algorithm, mobile manipulation for SFR, is shown, depicting the specific state space model of ExoTeR and the replanning methodology. 
In Section \ref{sec:results} the experimental results are depicted, including an analysis of the performance of the algorithm in the mobile manipulation for SFR use case and some laboratory tests with ExoTeR in the Planetary Robotics Laboratory (PRL).
In Section \ref{sec:discussion} the carried out experiments are examined to evaluate the strengths and weaknesses of the proposed algorithms.
Finally, Section \ref{sec:conclusions} concludes the paper with a final overview, including some comments on future related work.

\section{Multi-staged warm started Motion Planning} \label{sec:algorithm}
\begin{figure}[t]
    \centering
    \includegraphics[width=1\columnwidth]{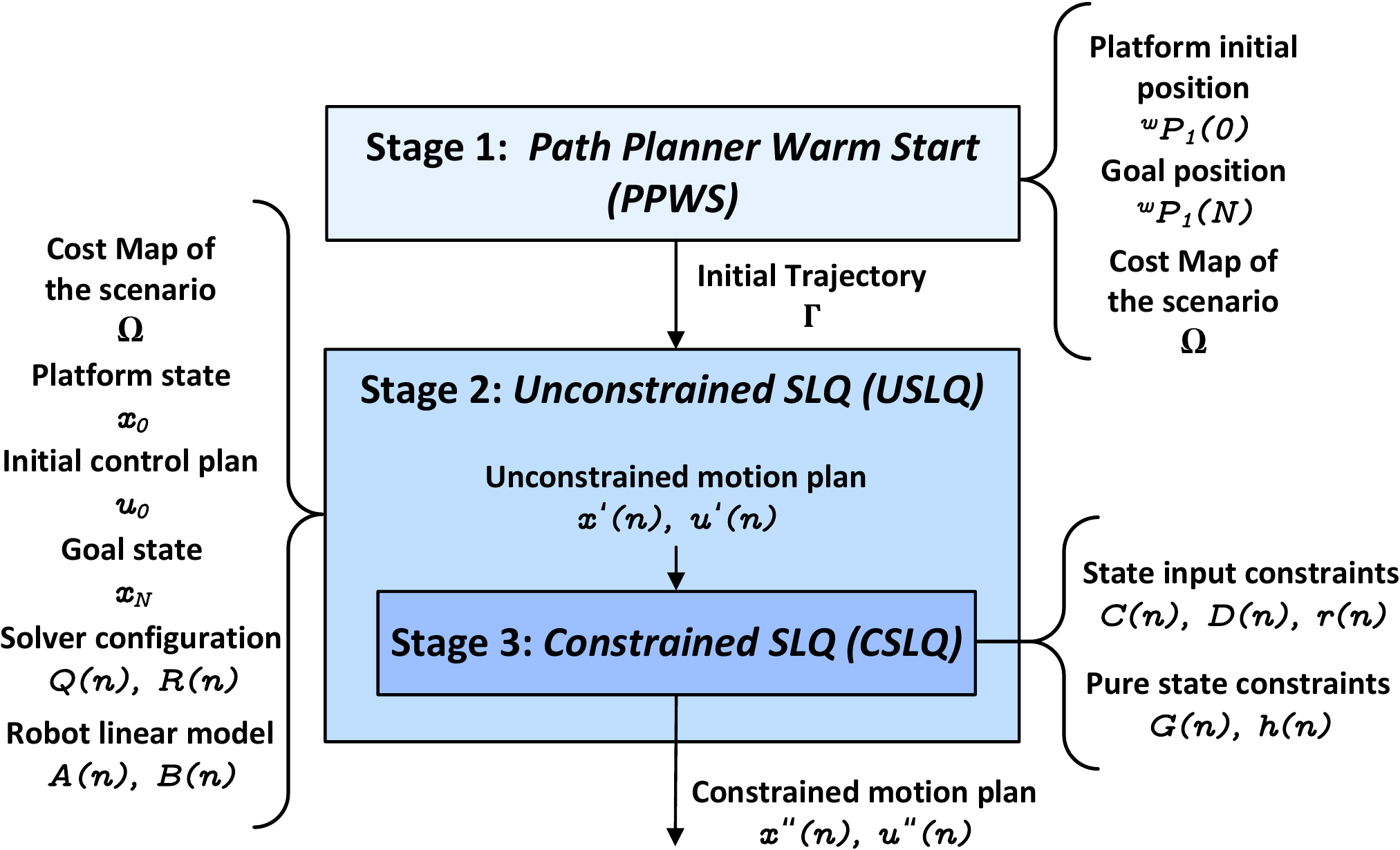}
    \caption{Scheme summarizing the general functioning of the multi-staged motion planning approach. In the first stage (PPWS) an initial trajectory $\Gamma$ is computed using Fast Marching Method (FMM). In the second stage (USLQ), this trajectory is used to warm start the Sequential Linear Quadratic (SLQ) optimization algorithm to solve the unconstrained motion planning problem. If the planned motion $x'(n), u'(n)$ satisfies the constraints, the algorithm finishes. Otherwise, the third stage (CSLQ) takes the unconstrained solution as warm start and computes the final motion plan $x''(n), u''(n)$, using again the SLQ optimization solver but with constraints compliance.}
    \label{fig:flowchart}
\end{figure}

The proposed Multi-staged Warm started Motion Planner (MWMP) is designed to deal with high complex, over-actuated systems, looking for an efficient solution of the motion planning problem without severely impacting the computational resources of the system. For that purpose, a sequential warm start procedure is defined in this section, which reduces substantially the average number of iterations until convergence and, thereupon, the computational cost of the planner.
A general overview of the functioning of the algorithm and its evolution through the different stages is shown in Figure \ref{fig:flowchart}, which is further explained below. 

Each stage is executed in a particular order, sequentially. 
The first stage, called Path Planning Warm Start (PPWS), consists of a path planner based on the Fast Marching Method (FMM), although any other path planning algorithm could be used. FMM has been selected since it provides globally optimal and smooth solutions, with comparable computational cost to other non-optimal state-of-the-art planners \cite{Sethian1999}.
As can be observed in Figure \ref{fig:flowchart}, the path planner requires three inputs to compute the trajectory: the platform initial and goal positions, $^wP_{1}(0)$ and $^wP_{1}(N)$ respectively, and a cost map $\Omega$ representing the characteristics of the scenario, with higher costs where the mobile platform finds difficulties to traverse, like obstacles.
Given this information, FMM takes two different steps to generate a path: first, a wave expansion, second, the trajectory extraction. This separation is advantageous, since it allows to generate new trajectories quickly as long as the goal remains the same, as will be explained later.
As an output, the path planner generates the global optimal path $\Gamma$ to reach the goal $^wP_{1}(N)$ from the platform initial pose $^wP_{1}(0)$ on the cost map $\Omega$. 

In order to accelerate the convergence speed, the extracted trajectory $\Gamma$ is forwarded as a warm start to the next stage, called Unconstrained SLQ (USLQ), which makes use of an optimal solver based on a Sequential Linear Quadratic (SLQ) regulator \cite{Sideris2005AnProblems}. 
Although many other optimal solvers could also be used, SLQ has been selected due to its efficiency when solving optimal control problems with near-quadratic convergence, besides, requiring only first derivative information of the system states.
As inputs, considering the set of $N$ time steps that define the planning horizon $T=\{t_0,\:t_1,\:...\:,\:t_n,\:...\:,\:t_N\}$, the second stage requires the current and goal states of the platform, $x_{0}$ and $x_{N}$, an initial control plan $u_0$ usually filled with zeros, a configuration for the solver i.e. the quadratic costs defined by the state matrix $Q(n)$ and the input matrix $R(n)$, and a linear state space model of the system, represented by the state transition matrix $A(n)$ and the input distribution matrix $B(n)$.
USLQ outputs, then, a complete unconstrained motion plan for the whole time horizon $T$, with $x'(n)$ the states and $u'(n)$ the actuation.

The above solver does not consider constraints yet, which implies faster convergence but could lead to unfeasible solutions. Thus, in the third stage, every generated solution is checked to confirm if the constraints are satisfied. If they are, then the unconstrained motion plan is taken as correct and the planning pipeline finishes, $x''(n) = x'(n)$ and $u''(n) = u'(n)$, being $x''(n)$ and $u''(n)$ the final motion plan states and actuation respectively. 
Otherwise, again the optimal solver is used to consider the system constraints, therefore, the third stage is called Constrained SLQ \cite{Sideris2011} (CSLQ). 
This stage is particularly critical for the feasibility of the motion plans but, besides, it is critical for the overall computational cost of MWMP, being the most computationally expensive stage \cite{Camacho2007ModelControl}. 
This is due to its intrinsic functioning: every time a constraint is violated (equality and inequality, state input and pure state constraints), the solver reduces proportionally the computed actuation to avoid that violation. 
Thus, the solver is encountering a barrier that reduces the number of alternatives where to explore the space of solutions. Even though CSLQ will eventually find a solution, it will take higher computational efforts, which is notably significant for mobile over-actuated systems: the variety of possible ways to reach a goal is enormous, and even more if we consider all the joints of the over-actuated system. 
Subsequently, the overall computational cost of MWMP is reduced noticeably if it avoids this constrained solver with a constraint-compliant solution found within the USLQ stage. 
Otherwise, when the constraints are violated, using the unconstrained solution as a warm start boosts noticeably the CSLQ performance, which is a main contribution of this work as will be shown with the obtained results in Section \ref{sec:use_case}. 
This way CSLQ only has to refine the unconstrained solution to ensure constraint compliance, i.e. it already starts in the vicinity of the constrained optimal solution. 

Therefore, the previously computed unconstrained motion plan $x'(n)$, $u'(n)$ is used to warm start the constrained solver, making it converge at much higher rates. 
Constrained SLQ additionally needs the definition of the constraints. 
On one hand, the system state input constraints, which indicate actuation caps under particular system states, for instance joint effort limits, defined by the state input constraints distribution matrices, with $C(n)$ for the state and $D(n)$ for the input, and the state-input constraints level $r(n)$. 
On the other hand, the pure state constraints, representing restrictions on the state vector that can not be overcome, for instance joint position limits, defined by the pure state constraints distribution matrix $G(n)$ and the pure state constraints level $h(n)$.

After finding a constraint-compliant solution, the complete motion plan $x''(n)$, $u''(n)$ is ready to be followed by the platform.
Any tracking or control algorithm could be used to accurately follow the planned motion.
For instance, MPC, NMPH, Event-triggered replanning or other procedures could be adequate for this purpose, in function of the platform and mission requirements.

Finally, the main prerequisite to use the motion planner is to obtain a state space model of the mobile platform, defining $A(n)$ and $B(n)$. Hence, a generic procedure for modelling over-actuated mobile platforms including several kinematic chains is presented below. 
Afterwards, the two main methods used in the motion planner, FMM as a path planner in the first stage and SLQ as an optimal solver for motion planning in the second and third stages, are analyzed with further details.

\subsection{State space model for a mobile platform}
Let us define a state space model under linear discrete approximations as depicted in (\ref{eq:state_space}):

\begin{equation}\label{eq:state_space}
    x(n+1) = A(n) x(n) + B(n) u(n)
\end{equation}

Where $A(n)$ and $B(n)$ are the state transition and input distribution matrices respectively, and $x(n)$ and $u(n)$ are the state and input vectors of the system respectively, at time step $n$. 
Assuming a generic mobile platform composed by a set of $K$ kinematic chains, its pose can be represented as $^jP_{k}$, which is the pose of the kinematic chain $k$ with respect to $j$, having $w$ as the world reference frame. 
Note that, in the following, any $^jP_{k}$ expresses the pose as $^jP_{k} = [p\:\phi]$, i.e. a position vector, in 3D $p = [x\:y\:z]$, and an orientation vector, in 3D $\phi = [\varphi  \:\vartheta \: \psi]$, the roll, pitch, yaw Euler Angles.
For instance, the center of a mobile platform would be $^wP_{1}$, and if it is equipped with a manipulator, the manipulator end effector pose w.r.t. the mobile platform would be $^1P_{2}$. 
The position of all the actuation joints belonging to kinematic chain $k$ are denoted as vector $q_k$, including rotational and traslational joints.
Thus, the corresponding state vector $x(n)$ of a generic mobile platform is defined as (\ref{eq:generic_x}).


\begin{equation} \label{eq:generic_x}
\begin{gathered}
x(n) = \bigl[ {^wP_{1}} \; ^w\dot{P}_{1}\; \gcdots \; ^wP_{K} \; ^w\dot{P}_{K} \; ^{1}P_{2} \; ^{1}\dot{P}_{2} \; \gcdots \; ^{K-1}P_{K} \\ \qquad \qquad  ^{K-1}\dot{P}_{K} \;
 q_{1} \; \dot{q}_{1} \; \ddot{q}_{1} \; q_{2} \; \dot{q}_{2} \; \ddot{q}_{2} \; \gcdots \; q_{K} \; \dot{q}_{K} \; \ddot{q}_{K} \bigr]^T
\end{gathered}
\end{equation}

With $^w\dot{P}_{k}$ the speed of the kinematic chain $k$ w.r.t. the world reference frame; $^{k-1}\dot{P}_{k}$ the speed of the kinematic chain $k$ w.r.t. the kinematic chain ${k-1}$; and $\dot{q}_{k}$, $\ddot{q}_{k}$ the speed and acceleration, respectively, of each actuation joint of $k$.
As a result, a generic representation of the state transition matrix $A(n)$ is shown in (\ref{eq:matrix_A}), where $\mathbb{I}_j$ represents the identity matrix with size $(jxj)$, $^jR_k$ a rotation matrix given the pose defined in $^jP_k$, $^{{j}}\mathcal{J}_{k}$ the Jacobian matrix relating articular with cartesian speeds for the kinematic chain $k$ w.r.t. ${j}$, $I_{k}$ and $V_{k}$ the inertia and Coriolis/centrifugal matrices of $k$ respectively, and $\Delta t$ the time step size.

\begin{table*}
\begin{equation}\label{eq:matrix_A}
A(n) = 
\begin{tikzpicture}[baseline=(current bounding box.center)]
\matrix[matrix of math nodes,execute at empty cell={\node[black!20]{\scriptstyle 0};},
        every left delimiter/.style={xshift=1ex},
        every right delimiter/.style={xshift=-1ex},
                left delimiter={[},right delimiter={]}
                ] {
            \mathbb{I}_6 &  & \gcdots &  &  &  &  & \gcdots &  &  &  & ^{w}\mathcal{J}_{1}\Delta t  &  &  &  &  & \gcdots &  &  &  \\
              &  & \gcdots &  &  &  &  & \gcdots &  &  &  & ^{w}\mathcal{J}_{1} &  &  &  &  & \gcdots &  &  &  \\
              \gvdots & \gvdots & \gddots & \gvdots & \gvdots & \gvdots & \gvdots & \gddots & \gvdots & \gvdots & \gvdots & \gvdots & \gvdots & \gvdots & \gvdots & \gvdots & \gddots & \gvdots & \gvdots & \gvdots \\
              &  & \gcdots & \mathbb{I}_6 &  &  &  & \gcdots &  &  &  &  &  &  &  &  & \gcdots &  & ^wR_{K-1}\: ^{K-1}\mathcal{J}_{K} \Delta t &  \\
              &  & \gcdots &  &  &  &  & \gcdots &  &  &  &  &  &  &  &  & \gcdots &  & ^wR_{K-1} \: ^{K-1}\mathcal{J}_{K} &  \\
              &  & \gcdots &  &  & \mathbb{I}_6 &  & \gcdots &  &  &  &  &  &  & ^{1}\mathcal{J}_{2} \Delta t &  & \gcdots &  &  &  \\
              &  & \gcdots &  &  &  &  & \gcdots &  &  &  &  &  &  & ^{1}\mathcal{J}_{2} &  & \gcdots &  &  &  \\
              \gvdots & \gvdots & \gddots & \gvdots & \gvdots & \gvdots & \gvdots & \gddots & \gvdots & \gvdots & \gvdots & \gvdots & \gvdots & \gvdots & \gvdots & \gvdots & \gddots & \gvdots & \gvdots & \gvdots \\
              &  & \gcdots &  &  &  &  & \gcdots & \mathbb{I}_6 &  &  &  &  &  &  &  & \gcdots &  & ^{K-1}\mathcal{J}_{K} \Delta t &  \\
              &  & \gcdots &  &  &  &  & \gcdots &  &  &  &  &  &  &  &  & \gcdots &  & ^{K-1}\mathcal{J}_{K} &  \\
              &  & \gcdots &  &  &  &  & \gcdots &  &  & \mathbb{I}_{q_1} & \mathbb{I}_{q_1}\Delta t &  &  &  &  & \gcdots &  &  &  \\
              &  & \gcdots &  &  &  &  & \gcdots &  &  &  & \mathbb{I}_{q_1} - I^{-1}_{1} V_{1} \Delta t &  &  &  &  & \gcdots &  &  &  \\
              &  & \gcdots &  &  &  &  & \gcdots &  &  &  & - I^{-1}_{1} V_{1} &  &  &  &  & \gcdots &  &  &  \\
              &  & \gcdots &  &  &  &  & \gcdots &  &  &  &  &  & \mathbb{I}_{q_2} & \mathbb{I}_{q_2} \Delta t &  & \gcdots &  &  &  \\
              &  & \gcdots &  &  &  &  & \gcdots &  &  &  &  &  &  & \mathbb{I}_{q_2} - I^{-1}_{2} V_{2} \Delta t &  & \gcdots &  &  &  \\
              &  & \gcdots &  &  &  &  & \gcdots &  &  &  &  &  &  & - I^{-1}_{2} V_{2} &  & \gcdots &  &  &  \\
              \gvdots & \gvdots & \gddots & \gvdots & \gvdots & \gvdots & \gvdots & \gddots & \gvdots & \gvdots & \gvdots & \gvdots & \gvdots & \gvdots & \gvdots & \gvdots & \gddots & \gvdots & \gvdots & \gvdots \\
              &  & \gcdots &  &  &  &  & \gcdots &  &  &  &  &  &  &  &  & \gcdots & \mathbb{I}_{q_K} & \mathbb{I}_{q_K} \Delta t &  \\
              &  & \gcdots &  &  &  &  & \gcdots &  &  &  &  &  &  &  &  & \gcdots &  & \mathbb{I}_{q_K} - I^{-1}_{K} V_{K} \Delta t &  \\
              &  & \gcdots &  &  &  &  & \gcdots &  &  &  &  &  &  &  &  & \gcdots &  & - I^{-1}_{K} V_{K} &  \\
        };
\end{tikzpicture}
\end{equation}
\end{table*}

On the other hand, assuming a force/torque controlled platform, the actuation vector $u(n)$ is defined as \ref{eq:generic_u}.

\begin{equation} \label{eq:generic_u}
    u = \left[
    \begin{gathered}
        e_{1} \:
        e_{2} \:
        \gcdots \:
        e_{K} \:\:
        \delta_1 \:
        \gcdots \:
        \delta_Z \:\:
    \end{gathered}
    \right]^T
\end{equation}

Where $e_{k}$ represents the actuation effort vector, i.e. forces or torques of the joints of $k$, and $\delta_z$ are external disturbances applying forces to the system, for instance, gravity.
Note that, even though the external disturbances $\delta_z$ are included in the model inside the control vector, doubtlessly they are not under control of the system, hence, they should remain fixed to their expected values.
The corresponding input distribution matrix $B(n)$ is depicted in \ref{eq:matrix_B}, with $\beta^k_z = -I^{-1}_{k} f^k_z$, and $f^k_z$ a matrix representing the effect of the perturbation $\delta_z$ into the joints of kinematic chain $k$.
It is crucial to remark that, with this definitions of $A(n)$ and $B(n)$, some of their terms will be non-linearities, which will eventually hinder the proper functioning of the motion planner. For that purpose, the use of Taylor Series Linearization (TSL) on those terms is recommended.


\begin{table}
\begin{equation}\label{eq:matrix_B}
B(n) = 
\begin{tikzpicture}[baseline=(current bounding box.center)]
\matrix[matrix of math nodes,execute at empty cell={\node[black!20]{\scriptstyle 0};},
        every left delimiter/.style={xshift=1ex},
        every right delimiter/.style={xshift=-1ex},
                left delimiter={[},right delimiter={]}
                ] {
             &  & \gcdots &  &  & \gcdots &  \\
             &  & \gcdots &  &  & \gcdots &  \\
             \gvdots & \gvdots & \gddots & \gvdots & \gvdots & \gddots & \gvdots \\
             &  & \gcdots &  &  & \gcdots &  \\
             &  & \gcdots &  &  & \gcdots &  \\
             &  & \gcdots &  &  & \gcdots &  \\
             &  & \gcdots &  &  & \gcdots &  \\
             \gvdots & \gvdots & \gddots & \gvdots & \gvdots & \gddots & \gvdots \\
             &  & \gcdots &  &  & \gcdots &  \\
             &  & \gcdots &  &  & \gcdots &  \\
             &  & \gcdots &  &  & \gcdots &  \\
            I^{-1}_{1} \Delta t  &  & \gcdots &  & \beta^1_1 \Delta t  & \gcdots & \beta^1_Z \Delta t  \\
            I^{-1}_{1} &  & \gcdots &  & \beta^1_1 & \gcdots & \beta^1_Z \\
             &  & \gcdots &  &  & \gcdots &  \\
             & I^{-1}_{2} \Delta t & \gcdots &  & \beta^2_1 \Delta t  & \gcdots & \beta^2_Z \Delta t \\
             & I^{-1}_{2} & \gcdots &  & \beta^2_1 & \gcdots & \beta^2_Z \\
             \gvdots & \gvdots & \gddots & \gvdots & \gvdots & \gddots & \gvdots \\
             &  & \gcdots &  &  & \gcdots &  \\
             &  & \gcdots & I^{-1}_{K} \Delta t & \beta^K_1 \Delta t  & \gcdots & \beta^K_Z \Delta t \\
             &  & \gcdots & I^{-1}_{K} & \beta^K_1 & \gcdots & \beta^K_Z \\
        };
\end{tikzpicture}
\end{equation}
\end{table}

Finally, it is key to define the system limits, via state-input and pure state constraints. On the one hand, the actuation effort $e_k$ limits have to be specified as state-input constraints, where $C(n)$ is filled with zeros, since this constraints do not depend on the states, $D(n)$ indicates with ones or zeros which actuation effort limit is being defined, and $r(n)$ includes the actual limit values. On the other hand, as pure state constraints it is important to define the kinematic chains world $^w\dot{P}_{k}$ and relative $^{k-1}\dot{P}_{k}$ speed limits, and the position ${q}_{k}$, velocity $\dot{q}_{k}$ and acceleration $\ddot{q}_{k}$ limits of the actuation joints. Particularly, $G(n)$ indicates which state limit is being defined, filled adequately with ones or zeros, and $h(n)$ includes the values of those limits.

Considering this general definition of $x(n)$,  $u(n)$, $A(n)$ and $B(n)$, the motion planning problem for a mobile platform with $K$ serial kinematic chains can be redefined as finding a set of actuation efforts (forces/torques) $e_{k}$ that generate a motion profile ($q_{k}$, $\dot{q}_{k}$, $\ddot{q}_{k}$) for each joint of the platform, in order to place the last kinematic chain in certain poses ($^wP_{K}$, $^w\dot{P}_{K}$), given the effect of external perturbations ($\delta_j$) and the system limits, expressed through the state-input constraints ($C(n)$, $D(n)$ and $r(n)$) and the pure state constraints ($G(n)$ and $h(n)$).
Note that some of the states are not strictly necessary in the model ($^w\dot{P}_{k}$, $^{k-1}\dot{P}_{k}$, $\ddot{q}_{k}$), but, as will be explained later, are helpful to set desired behaviours by tuning their corresponding costs, or to establish system constraints as aforementioned.


\subsection{Trajectory planning with FMM}
The goal of the first stage, PPWS, is to generate an initial reference trajectory for the mobile platform to reach the goal, which will be later used as a warm start of the optimization algorithm to accelerate its convergence, as can be seen in Figure \ref{fig:flowchart}. 
In particular, we propose Fast Marching Method (FMM) \cite{Sethian1999} as the warm start path planner, since, considering the scenario in form of a cost map, it extracts a globally optimal, smooth and continuous path to reach the goal.
Therefore, the FMM warm start induces the mobile platform to follow the optimal Cartesian 2D/3D space trajectory, making MWMP a partially optimal approach.
Remark that FMM has been widely used in the literature as a path planner \cite{Valero-Gomez2013} for different applications, that span from Unmaned Surface Vehicles (USVs) \cite{Liu2015} to planetary rovers \cite{Sanchez-Ibanez2019}. 

First of all, as aforementioned, FMM requires a proper representation of the scenario as an input cost map $\Omega$. 
The cost map $\Omega$ is a discrete 2D or 3D grid, where each regularly scattered node $\Tilde{x}_{j}$ has an associated cost $\Omega(\Tilde{x}_{j})$ that represents how easy and safe is for the platform to be placed in that position. 
Subsequently, obstacles should have the highest costs, and traversable areas the lowest ones. Areas surrounding obstacles should also have high costs, to avoid the platform getting close to them. 
Additionally, any other feature that influences the platform behaviour should be considered in the cost map. For instance, slopes and terramechanic properties of the soil in the case of Unmanned Ground Vehicles (UGVs).

FMM numerically solves a particular non linear Partial Derivative Equation (PDE) called the Eikonal equation, modelling the rate of propagation of a wave.
This wave expands on the cost map $\Omega$ from the goal node $\Tilde{x}_{g}$ visiting each node $\Tilde{x}_{j}$ to generate the cost to go $\Upsilon(\Tilde{x}_{j}, \Tilde{x}_{g})$, which indicates the accumulation of cost required to reach the goal from the node $\Tilde{x}_{j}$.
As indicated in (\ref{eq:eikonal}), the rate of propagation of the wave at a certain node is equal to the cost at that node $\Omega(\Tilde{x}_{j})$. 


\begin{equation} \label{eq:eikonal}
    ||\nabla \Upsilon(\Tilde{x}_{j}, \Tilde{x}_{g})|| = \Omega(\Tilde{x}_{j}) \ \ \ \forall \Tilde{x}_{j} \in \Omega
\end{equation}

The higher the cost $\Omega(\Tilde{x}_{j})$ the slower the propagation of the wave on that node $\Tilde{x}_{j}$. 
In this case, the wave propagation starts from the goal $\tilde{x}_{g}$, therefore the cost to go of the goal node is zero ($\Upsilon(\Tilde{x}_{j} = \Tilde{x}_{g},  \Tilde{x}_{g}) = 0$).

The cost to go between the starting  $\Tilde{x}_{0}$ and the goal nodes $\Tilde{x}_{g}$ is the minimum possible if $\Omega(\Tilde{x}_{j})$ always returns positive non-zero values.
Thus, following the Dynamic Programming Principle (DDP), any point $\hat{x} \in \Omega$ is placed in the optimal path connecting the starting and goal nodes, $\Gamma(\tilde{x}_{0},\tilde{x}_{g})$, if the sum of the costs to go from the starting node to the point $\Upsilon(\Tilde{x}_{0}, \hat{x})$ and from the point to the goal node $\Upsilon(\hat{x}, \Tilde{x}_{g})$ is equal to the minimum cost to go $\Upsilon(\Tilde{x}_{0}, \Tilde{x}_{g})$, as expressed in (\ref{eq:dp_principle}).


\begin{equation}\label{eq:dp_principle}
    \Upsilon(\tilde{x}_{0}, \hat{x}) + \Upsilon(\hat{x},\tilde{x}_{g}) = \Upsilon(\tilde{x}_{0},\tilde{x}_{g}) \ \ \ \forall \hat{x} \in \Gamma(\tilde{x}_{0},\tilde{x}_{g}) \in \Omega
\end{equation}

Hence, the objective of FMM is to solve the optimization problem defined in (\ref{eq:totalcostcase2}-\ref{eq:totalcost_goalvalue}), i.e finding the optimal path $\Gamma(\tilde{x}_{0},\tilde{x}_{g})$ that minimizes the cost accumulated along the path $\Omega(\Gamma(\Tilde{x}_{0}, \Tilde{x}_{g}, v))$, being $\Gamma(\Tilde{x}_{0}, \Tilde{x}_{g}, v)$ a continuous function that returns a point $\hat{x} \in \Omega$ given the path length $v$ from the starting node $\Tilde{x}_{0}$.

\begin{equation}\label{eq:totalcostcase2}
    \begin{split}
        \textmd{Minimize} \\
        {\Gamma(\Tilde{x}_{0}, \Tilde{x}_{g})}
    \end{split}
    \quad \Upsilon(\Tilde{x}_{j}, \Tilde{x}_{g}) = \int_{v_{j}}^{v_g} \Omega(\Gamma(\Tilde{x}_{0}, \Tilde{x}_{g}, v))\:dv 
\end{equation}

\begin{equation}\label{eq:totalcost_goalvalue}
    \textmd{with} \quad \Upsilon(\Tilde{x}_{j} = \Tilde{x}_{g},  \Tilde{x}_{g}) = 0
\end{equation}

Considering $1$ as the kinematic chain defining the mobile platform, then $\Tilde{x}_{0} = {^wP_{1}(0)}$ and $\Tilde{x}_{g} = {^wP_{1}(N)}$.
Consequently, the warm start trajectory $\Gamma$ is used as a reference for the pose of the platform $^wP_{1}$ at each time step $n$.

The wave propagation is, for FMM, the most computationally expensive step, being the trajectory extraction computationally negligible. This is very convenient for replanning the motion, since a new optimal trajectory from the current pose of the platform can be obtained quickly, i.e. without recomputing the cost to go. This only needs to be done once, offline, or in case that the goal changes.
Refer to \cite{Sanchez-Ibanez2019} for more details about the FMM based path planner.

The generated path, nevertheless, should be enriched with some more information to become a motion plan. 
On one hand, the orientation of the platform at each waypoint is computed to also warm start the orientation states of the system. This is particularly helpful for platforms with non-holonomic constraints, which is usually the case of UGVs.
Including the yaw in the trajectory gives the mobile platform a big hint on how to properly follow the path, being the yaw obtained geometrically known the position of two consecutive waypoints.
On the other hand, each waypoint should be timestamped, which can be easily done by interpolating the total expected time for finishing the operation $t_N$ at each time step $n$. 
Note that there are many different approaches to estimate $t_N$ according to the characteristics of the system, its nominal speed or the use case, which is out of the scope of this paper.

\subsection{Sequential Linear Quadratic (SLQ) optimal solver}
The next stages of MWMP make use of an optimal solver, which tackles the unconstrained problem in stage two (USLQ) and the constrained one in stage three (CSLQ), to generate a motion plan for the platform, as shown in Figure \ref{fig:flowchart}. 
The solver is called Sequential Linear Quadratic (SLQ) algorithm \cite{Sideris2005AnProblems}\cite{Sideris2011}, and, based on the Riccati equation, it is characterized by efficiently solving non-linear discrete optimal control problems with near-quadratic convergence. 

Considering the set of $N$ time steps that define the planning horizon $T=\{t_0,\:t_1,\:...\:,\:t_n,\:...\:,\:t_N\}$, the standard formulation of a discrete-time optimal control problem is shown in (\ref{eq:optimal_control_definition}-\ref{eq:ps_constraints}):

\begin{equation}\label{eq:optimal_control_definition}
    \begin{split}
        \textmd{Minimize} \\
        u(n), x(n)
    \end{split}
    \quad J = \Phi(x(N)) + \sum_{n=0}^{N-1}L(x(n),u(n),n)
\end{equation}

\begin{equation}\label{eq:model_and_initial_state}
    \textmd{subject to} \quad x(n + 1) = f(x(n),u(n)), \quad x(0) = x_0
\end{equation}
\begin{equation}\label{eq:si_constraints}
    \quad\quad C(n)x(n) + D(n)u(n) + r(n) \leq 0 \:
\end{equation}
\begin{equation}\label{eq:ps_constraints}
    \:G(n)x(n) + h(n) \leq 0 \quad\quad\quad\quad
\end{equation}

Where $x(n)$ is the state vector at time-step $n$, noticeably, $x(0)$ is the initial state and $x(N)$ is the final one, and $u(n)$ is the actuation vector; $x_0$ defines the initial state of the system, at time step $t_0=0$.
Additionally, (\ref{eq:si_constraints}) represents the state input inequality constraints with the constraints distribution matrices $C(n)$, $D(n)$ and the constraints level vector $r(n)$, and (\ref{eq:ps_constraints}) the pure state constraints with the constraints distribution matrix $G(n)$ and the constraints level vector $h(n)$, as aforementioned. 

In this formulation, $J$ is defined as the total cost to go, and it is composed of $\Phi(x(N))$, the terminal cost, and  $L(x(n),u(n),n)$, the intermediate cost. Assuming a quadratic performance index, these are defined in (\ref{eq:terminal_cost}) and (\ref{eq:intermediate_cost}) respectively:

\begin{equation}\label{eq:terminal_cost}
    \Phi(x(N)) = \frac{1}{2}[x(N)-x^0(N)]^T Q(N) [x(N)-x^0(N)]
\end{equation}

\begin{equation}\label{eq:intermediate_cost}
\begin{aligned}
    L(x(n),u&(n),n) = \\
    \frac{1}{2}&[x(n) - x^0(n)]^T Q(n)[x(n)-x^0(n)] \\
          + &[u(n) - u^0(n)]^T R(n)[u(n)-u^0(n)]
\end{aligned}
\end{equation}

With $Q(n)$, $R(n)$ defined as the state and input quadratic cost matrices respectively, at time step $n$, and $x^0(n)$, $u^0(n)$ the state and input references or targets. Note that $x^0(N) = x_N$ is the terminal state goal and $Q(N)$ the terminal state cost matrix, which is usually configured to have considerably high costs to ensure that the goal is accomplished. 
Note also the importance of properly tuning $Q(n)$ and $R(n)$ to precisely represent the desired behaviour of the system.
In fact, this costs can be dynamically tuned during the execution of the motion plan depending on how the system is evolving, to achieve newly desired behaviours or to emphasize them. 

Solving the aforementioned discrete-time optimal control problem, the objective of the algorithm is to generate the motion plan, $x$ and $u$, for the whole time horizon $T$. 
To that purpose, several inputs are required. On one hand, the current state of the system $x_0$ and the desired goal state $x_N$ are needed. On the other hand, as extensively explained above, an initial trajectory $\Gamma$ is fed to the solver to accelerate the convergence speed. 
Consequently, the corresponding intermediate state costs $Q(^wP_{1}, n)$ must be tuned with appropriate costs at every time step. 
These have to be high enough to help the solver, guiding it closer to the globally optimal path, but low enough to avoid forcing the solver to follow exactly the provided trajectory, which would reduce the variety of possible solutions to be explored.

Additionally, obstacle avoidance is always a requirement for any mobile platform. Although the warm start trajectory already considers the presence of obstacles in the scenario, another layer of obstacle avoidance is required, since the solver will draw a probably similar but new trajectory. 
Thus, USLQ and CSLQ need the same cost map $\Omega$ used in FMM at PPWS. $\Omega(^wP_{1}(n))$ corresponds to a repulsive cost that gets the system away from danger, which means that the cost increases as the system gets closer to obstacles.
It is key that the continuity and linearity of the cost map are maximized, otherwise the solver will find difficulties to converge when encountering non-linearities in the costs. 

The generated trajectories for the platform base pose $^wP_{1}$ dynamically get away from obstacles during the motion planning process in function of the cost map $\Omega$, meanwhile trying to follow the warm start trajectory $\Gamma$. The intermediate cost defined above in (\ref{eq:intermediate_cost}) needs, then, to be reformulated, adding the repulsive cost $\Omega(^wP_{1}(n))$, i.e. the cost value associated to the platform pose $^wP_{1}$ at time step $n$.


\begin{algorithm}
    \begin{algorithmic}[1]
        \STATE \textbf{Initialization}
        \STATE $x(0) \leftarrow$ getCurrentState()
        \STATE $u \leftarrow$ getCurrentControlPlan()
        \STATE $x^0(^wP_{1}) \leftarrow$ getWarmStartTrajectory()
        \STATE $C, D, r, G, h \leftarrow$ getConstraints()
        \REPEAT
            \STATE \textbf{Linearization and quadratization}
            \STATE $x \leftarrow$ forwardSimulateSystem($x(0)$, $u$)
            \STATE $Q, R \leftarrow$ getQuadraticCosts()
            \STATE $A, B \leftarrow$ getLinearizedSystem($x$)
            \STATE $\Omega \leftarrow$ getObstaclesRepulsiveCost($F(x)$)
            \STATE $C_c, D_c, G_c \leftarrow$ getActiveConstraints($C, D, r, G, h, x, u$)
            
            \STATE \textbf{Reference tracking}
            \STATE $\bar{x}_0 \leftarrow$ $Q(x - x^0)$
            \STATE $\bar{u}_0 \leftarrow$ $R(u - u^0)$
            
            \STATE \textbf{Predefinitions}
            \STATE $\hat{D} \leftarrow (D_c R^{-1}D_c^T)^{-1}$
            \STATE $\hat{r} \leftarrow - D_c R^{-1} \bar{u}_0$
            \STATE $\hat{A} \leftarrow A - B R^{-1} D_c^T \hat{D} C_c$
            \STATE $\hat{R} \leftarrow B R^{-1} [\mathbb{I} - D_c^T \hat{D} D_c R^{-1}] B^T$
            \STATE $\hat{Q} \leftarrow Q + C_c^T \hat{D} C_c$
            \STATE $\hat{x}^0 \leftarrow \bar{x}_0 + C_c^T \hat{D} \hat{r}$
            \STATE $\hat{u}^0 \leftarrow -B R^{-1} [\bar{u}_0 + D_c^T \hat{D} \hat{r}]$

            \STATE \textbf{Backward Pass - Riccati matrix difference equation}
                \STATE $\hat{P}(N) \leftarrow Q(N)$
                \STATE $s(N) \leftarrow \Omega(N) + \bar{x}_0$
                \FOR{$n \leftarrow (N-1)$\textbf{;} $n$ \textbf{in} $T$}
                    \STATE $\hat{M}(n) \leftarrow (I + \hat{R}(n) \hat{P}(n+1))^{-1}$
                    \STATE $\hat{P}(n) \leftarrow \hat{Q}(n) + \hat{A}^T(n) \hat{P}(n+1) \hat{M}(n) \hat{A}(n)$
                    \STATE $s(n) \leftarrow \hat{A}^T(n) \hat{M}(n)^T s(n+1) +$ \\ \quad $ \hat{A}^T(n) \hat{P}(n+1) \hat{M}(n) \hat{u}^0(n) + \hat{x}^0(n) + \Omega(n)$
                \ENDFOR
            \STATE \textbf{State constraints management} (ref to Alg \ref{alg:state_constraints})
            
            \STATE \textbf{Forward Pass}
                \FOR{$n \leftarrow 0$\textbf{;} $n$ \textbf{in} $T$}
                    \STATE $\hat{v}(n) \leftarrow \hat{M}(n) (\hat{u}^0(n) - \hat{R}(n) s(n+1))$
                    \STATE $\bar{x}(n+1) \leftarrow \hat{v}(n) + \hat{M}(n) \hat{A}(n) \bar{x}(n)$
                    \STATE $\lambda(n+1) \leftarrow s(n+1) + \hat{P}(n+1) \bar{x}(n+1)$
                    \STATE $\mu(n) \leftarrow \hat{D}(n) [C_c(n) \bar{x}(n) - $ \\ \quad $D_c(n) R^{-1}(n) B^T \lambda(n+1)  + \hat{r}(n)]$

                    \STATE $\bar{u}(n) \leftarrow - R^{-1}(n) [ B^T(n) \lambda(n+1) + $ \\ \quad $D_c(n)^T \mu(n) + \bar{u}_0(n)]$
                \ENDFOR
            \STATE \textbf{Computed step plan appliance}
            \IF{checkConstraints($x, \bar{x}, u, \bar{u}, C, D, r, G, h$) \textbf{or} \\ \quad isUnconstrained()}
                \STATE $\alpha \leftarrow$ computeLineSearch($x, x^0, u, u^0, \bar{u}$)
            \ELSE
                \STATE $\alpha \leftarrow$ satisfyConstraints($x, \bar{x}, u, \bar{u}, C, D, r, G, h$)
            \ENDIF
            \STATE $x \leftarrow x + \alpha \bar{x}$
            \STATE $u \leftarrow u + \alpha \bar{u}$

            \STATE \textbf{Termination conditions}
            \STATE convergence $\leftarrow$ checkTermination($x, x^0, u, u^0, \bar{u}, \alpha$)

        \UNTIL{convergence}
    \end{algorithmic}
    \caption{SLQ solver}
    \label{alg:constrained_slq}
\end{algorithm}

\begin{algorithm}
    \begin{algorithmic}[1]
        \IF{\textbf{not} isUnconstrained()}
            \STATE \textbf{Solve over all state constraints $c$}    
            \FOR{$c \in \mathcal{C}_x = {c_1, ..., c_S}$}
                \STATE $\Psi_c(t_c) \leftarrow G_c(t_c)$
                \STATE $y_c(t_c) \leftarrow 0$
                \FOR{$n \leftarrow (t_c-1); n--; n>0$}
                    \STATE $\Psi_c(n) \leftarrow \Psi_c(n+1) \hat{M}(n) \hat{A}(n)$
                    \STATE $y_c(n) \leftarrow y_c(n+1) + $\\ \quad $ \Psi_c(n+1) \hat{M}(n) [\hat{u}^0(n) - \hat{R}(n) z(n+1)]$
                \ENDFOR
                \STATE $H(c) \leftarrow h_c(t_c)$
                \STATE $\Psi(c) \leftarrow \Psi_c(0)$
                \STATE $y(c) \leftarrow y_c(0)$
                
                \FOR{$j \in \mathcal{C}_x = {j_1, ..., j_S}$}
                    \STATE $n \leftarrow min(c-1, j-1)$
                    \STATE $F_{c,j}(n+1) \leftarrow 0$
                    \FOR {$i \leftarrow n;  i\geq0$}
                        \STATE $F_{c,j}(i) \leftarrow F_{c,j}(i+1) - $\\ \quad $\Psi_c(i+1) \hat{M}(i) \hat{R}(i)\Psi_j(i+1)^T$
                    \ENDFOR
                    \STATE $F(c, j) \leftarrow F_{c,j}(0)$
                \ENDFOR
            \ENDFOR
            \STATE \textbf{Compute the Lagrangian Multipliers}
            \STATE $\nu \leftarrow F^{-1}[-\Psi \bar{x} + y + H]$
            \FOR{$n \leftarrow 0, ..., N$}
                \STATE $s(n) \leftarrow s(n) + \sum_{c\in \mathcal{C}_x; c\geq n}^{} \Psi_c^T(n) \nu(c)$
            \ENDFOR
        \ENDIF
    \end{algorithmic}
    \caption{State constraints management}
    \label{alg:state_constraints}
\end{algorithm}

Finally, an overview of the functioning of the solver for the discrete-time optimal control problem defined in (\ref{eq:optimal_control_definition}-\ref{eq:ps_constraints}) is depicted in Algorithm \ref{alg:constrained_slq}. 
Summarizing, given the current state $x_0$, the current control plan $u$, the warm start trajectory $\Gamma$, the quadratic costs $Q$ and $R$, the system model $A$ and $B$, the cost map of the scenario $\Omega$, the state input $C, D, r$ and the pure state $G, h$ constraints, this solver computes efficiently the motion plan $x, u$ by iteratively obtaining step plans $\bar{x}, \bar{u}$ to be applied to the current solution. 
To do so, the current active constraints are stored in $C_c$, $D_c$ and $G_c$, which are later used to consider the constraints during the LQR solution computation. 
In particular, the state-input constraints are directly handled within the \textit{Predefinitions} step, meanwhile the pure state constraints are managed later as detailed in Algorithm \ref{alg:state_constraints}.

Algorithm \ref{alg:constrained_slq} is based on the SLQ solver presented in \cite{Sideris2005AnProblems} and \cite{Sideris2011}, with a few differences. 
First, the state target sequence $x^0(n)$ is initialized with the warm start trajectory, as aforementioned. 
Second, the obstacles repulsive cost $\Omega(n)$ is included into the backward pass. 
Third, at each iteration the constraints compliance is checked. If no new constraint is violated, then a standard \textit{Line Search} for $\alpha$ is performed. 
The line search consists in finding the best $\alpha$, which is the step size used to apply the step solutions $\bar{x}$ and $\bar{u}$, in order to reduce to the minimum the total cost to go $J$ at each iteration of the solver.
Otherwise, if any constraint is violated, $\alpha$ is generated particularly to satisfy the constraints.
Fourth, several termination conditions are defined, on top of the algorithm convergence itself.
In particular, one of these conditions checks that the last kinematic chain pose $^wP_{K}$ is close enough to the goal pose $^wP_{K}(N)$ to perform the desired task, depending on a given threshold, and another one ensures that the motion plan is thoroughly safe.
Note that Algorithm \ref{alg:constrained_slq} encompasses both the unconstrained and the constrained solvers, with $[C, D, r, G, h, C_c, D_c, G_c] = 0$ in the unconstrained case, which means that $\hat{A} = A$, $\hat{R} = B R^{-1} B^T$, $\hat{Q} = Q$, $\hat{x}^0 = \bar{x}_0$, $\hat{u}^0 = -B R^{-1} \bar{u}_0$ and $\mu(n) = 0$. Besides, as can be observed, for the unconstrained case the \textit{State constraints management} is not required, and the \textit{Computed step plan appliance} is reduced to the first \textit{Line Search}.

\section{Use case: Sample Fetch Rover} \label{sec:use_case}

Planetary exploration vehicles are requiring more and more autonomy  since remote teleoperation from Earth hinders to perform complex tasks such as navigation and manipulation \cite{Bajracharya2008AutonomyFuture}. 
A common strategy to increase autonomy for mid-long range traverses in planetary surfaces is a Guidance, Navigation and Control (GNC) architecture \cite{Gerdes2020EfficientResources}, which allows the rover to plan a path to the goal and navigate safely to it avoiding any intermediate hazard. 
Nevertheless, future Mars vehicles like the Sample Fetch Rover (SFR) \cite{Merlo2013SampleMSR} demand a further effort on the autonomous capabilities of the system, to satisfy the time constraints imposed by the mission. SFR will collect several soil sample tubes, left by the Perseverance Mars2020 rover, to eventually bring them back to the Earth, with the requirement of going to the samples location, retrieving the samples and coming back to the lander within 150 sols \cite{Muirhead2019MarsConcepts}.
Considering this critical time restriction, the necessity of performing the sample retrieval operations autonomously and efficiently arises, to increase the overall navigation speed of the system.

SFR will be a highly over-actuated mobile platform, composed of a mobile base with multiple actuators (mainly driving and steering joints), and a robotic arm with several DoF, to retrieve the sample tubes. 
Considering, besides, the energy and time efficiency requirements, it is the perfect use case to demonstrate the advantages of the proposed optimal motion planning methodology.

In particular, a prototype of the Rosalind Franklin ExoMars rover from the Planetary Robotics Laboratory of the European Space Agency (ESA-PRL), called ExoTeR (Exomars Testing Rover) \cite{Azkarate2022DesignExploration}, will be used. It is a triple-bogie, non-holonomic and double-Ackermann steered rover, equipped with a 5 DoF manipulator, which is modelled following the aforementioned generic over-actuated mobile platform state space model.
Besides, in order to test the motion planner within a real analogue SFR mission, it is necessary to wrap it into a motion plan follower. 
This follower will filter any external disturbances or errors during the execution of the motion plan, by replanning the motion when significant deviations are measured. 

Thus, an analysis on the platform characteristics, a depiction on the developed state space model of the mobile manipulator and a detailed description of the tailored replanning methodology are presented in this section.

\subsection{Mobile platform description}

\begin{figure}[t]
    \centering
    \includegraphics[width=\columnwidth]{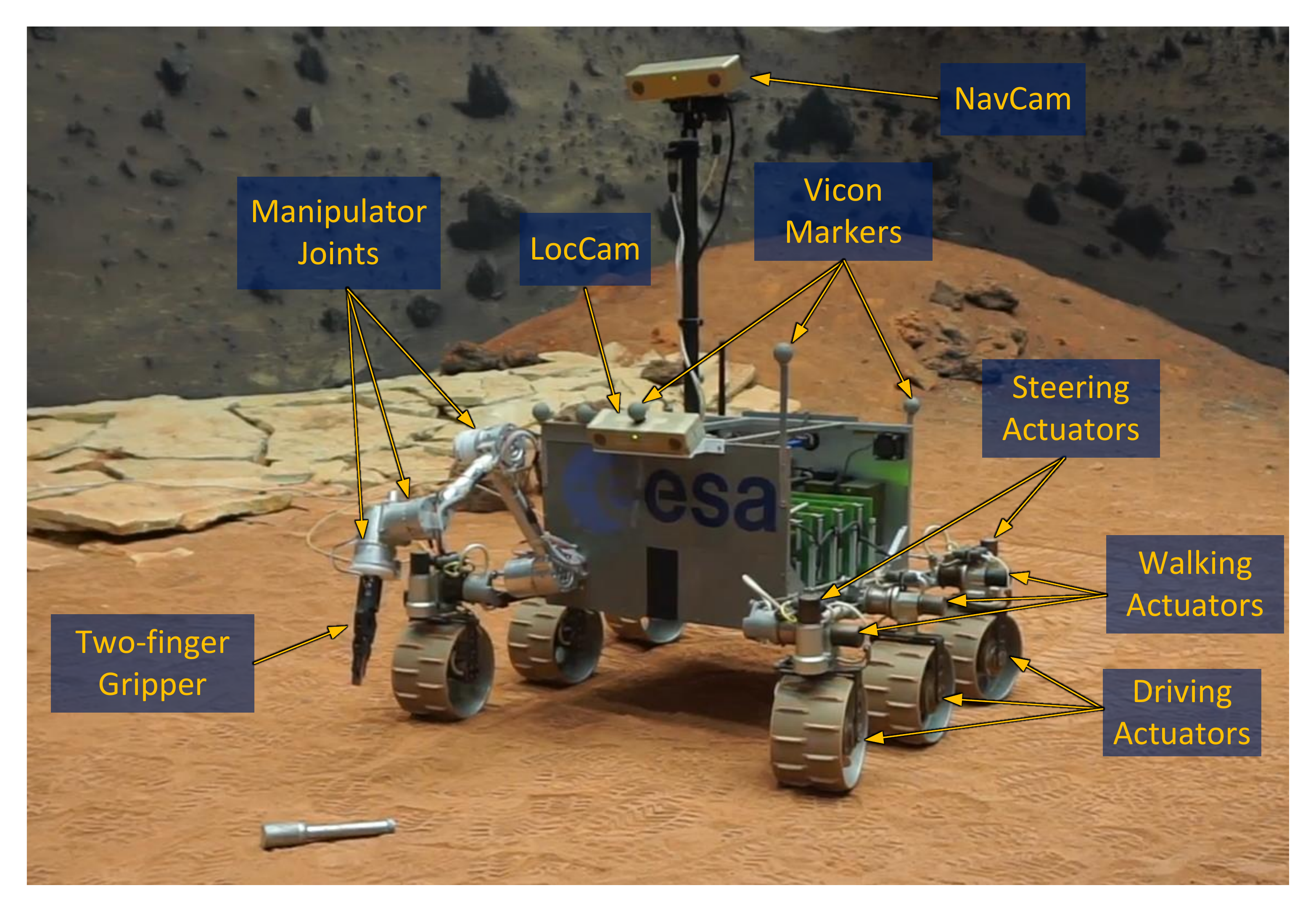}
    \caption{Detail of the experimental setup with ExoTeR approaching a sample tube, including its actuators (driving, walking, steering and manipulator joints, two-fingers gripper) and its exteroceptive sensors (LocCam, NavCam). Besides, ExoTeR is equipped with an Inertial Measurement Unit (IMU) to estimate its orientation and several Vicon Markers to precisely locate it inside the Martian Analogue Testbed of the Planetary Robotics Laboratory, ESA-ESTEC.}
    \label{fig:exoter_depiction}
\end{figure}

Within the research and development carried out at the Planetary Robotics Laboratory, Automation and Robotics  Section, of the European Space Agency (ESA-PRL), the design and testing of planetary rover testbeds stands out. 
This is the case of ExoTeR \cite{Azkarate2022DesignExploration}, which conceptually mimics the early model of the Rosalind Franklin ExoMars rover, with a scaled-down concept. ExoTeR is a triple-bogie, double-Ackermann rover, with a locomotion system of 6 x 6 x 4 + 6. 
This means 6 wheels with 6 driving actuators, 4 of them steerable (the front and rear ones), which permits double-Ackermann steering or spot turns. 
Additionally, all 6 wheels include a walking actuator, as depicted in figure \ref{fig:exoter_depiction}, where ExoTeR is shown at the Martian Analogue Testbed at ESA-PRL. 
ExoTeR is also equipped with a 5 DoF manipulator, called MA5-E. Its five joints are rotational, with a Roll-Pitch-Pitch-Pitch-Roll configuration, being the first joint placed looking towards the movement direction of the platform. 
Its end effector has attached a two-fingered gripper for sample retrieval purposes. 
For localization and perception, ExoTeR has two stereo cameras, a close range LocCam and a long range NavCam. 
Finally, ExoTeR has also an Inertial Measurement Unit (IMU) for sensing the platform 3D orientation, and has appended several Vicon markers for ground-truth localization inside the martian testbed. 

Focusing first on the mobile platform base, ExoTeR is a double-Ackermann rover with steering joints at the front and rear wheels. 
The central wheels do not steer, which implies that the rover can not move in every direction depending on the system orientation, i.e. non-holonomic constraints. This is a significant non-linearity, which is tackled inside the system model. 
The platform minimum turn radius is $0.6 m$, due to the geometric distribution of the wheels and the range limit of $\pm50^{\circ}$ in the steering joints, although it can perform point turns (change its orientation with zero linear velocity).
Finally, its nominal traslational speed is $5 cm/s$, with a $2.85 N m$ maximum torque of the driving actuators.

Regarding the robotic arm, MA5-E, its main characteristics are outlined in Table \ref{tab:MA5-E_charac}.
As can be observed, MA5-E joints have huge gear ratios, which allows it to handle heavy payloads, $2 kg$, in comparison to the arm weight, $2.4 kg$, and considering the joint motors power ($6 W$).
The dynamic effect of external disturbances is consequently negligible, i.e. gravity and the rover base movements.
Nevertheless, the gears also imply an important drawback: the joints move very slowly, with a maximum rotational speed of $0.57^\circ/s$.

\begin{table}[t]
    \centering
    \caption{MA5-E joints characteristics.}
    \vspace{3mm}
    \begin{tabular}{| c | c | c | c | c | c |}
        \hline
        Joint & 1 & 2 & 3 & 4 & 5 \\
        \hline
        Type & Rot. & Rot. & Rot. & Rot. & Rot.\\
        \hline
        Orientation & Z & Y & Y & Y & Z \\
        \hline
        Range ($^\circ$)& $\pm 45$ & $\pm 170$ & $\pm 170$ & $\pm 170$ & $\pm 170$ \\
        \hline
        Speed ($^\circ/s$) & $0.57$ & $0.57$ & $0.57$ & $0.57$ & $0.57$ \\
        \hline
        Power ($W$) & $0.75$ & $0.75$  & $0.75$  & $0.75$  & $0.75$ \\
        \hline
        Gear ratio & 83200:1 & 83200:1 & 83200:1 & 83200:1 & 83200:1 \\
        \hline
        Efficiency & $0.5$ & $0.5$ & $0.5$ & $0.5$ & $0.5$ \\
        \hline
    \end{tabular} 
    \label{tab:MA5-E_charac}
\end{table}

The fully extended arm lengths $0.527m$, with additional $0.14m$ taking into account the gripper. The arm end effector reachability is restricted by each joint position limit, as can be observed in Table \ref{tab:MA5-E_charac}.
Doubtlessly, the arm movements are also limited by the rover body itself and the ground.
Regarding the end effector, it can not reach any orientation because of the limitation of the 5 DoF configuration. This issue hinders any manipulation task, especially a sample retrieval operation, since the end effector could not approach the sample completely perpendicularly to the ground with the appropriate gripper yaw w.r.t. the sample tube.


Nonetheless, the optimal motion planner can tackle easily the manipulator constraints.
First, a coupled arm-base motion solves the arm joints velocity issues, generating optimal motions where the arm is already prepared to perform the desired task once the base has reached the objective. 
Second, the rover DoFs can be used to place the manipulator in a certain manner to effectively retrieve the sample, i.e. aligning the manipulator first joint and the sample tube, performing completely perpendicular retrieval operations. This is achieved by properly tuning the costs associated to the goal pose of the end effector, including its orientation, as will be clarified in Section \ref{sec:results}.

\subsection{Double-Ackermann mobile manipulator model}
As a mobile manipulator, ExoTeR is modelled with two different kinematic chains: the full-Ackermann mobile base and the robotic arm.
The dynamics coupling between them is ignored, seeing that the movements of both the platform and the manipulator are very slow, generating negligible dynamics effects between them.
Additionally, the effect of gravity into the manipulator joints is also disregarded, considering the huge gear ratios as aforementioned. 

Following the generic state space model explained before, the state vector $x(n)$ for ExoTeR is defined as follows:

\begin{equation}
    x(n) = \left[
    \begin{gathered}
        ^wP_{1} \; 
        ^w\dot{P}_{1} \; 
        ^wP_{2} \; 
        ^{1}P_{2} \; 
        ^{1}\dot{P}_{2} \; 
        q_{1} \; 
        \dot{q}_{1} \; 
        \ddot{q}_{1} \; 
        q_{2} \; 
        \dot{q}_{2} \; 
        \ddot{q}_{2}
    \end{gathered}
    \right]^T
\end{equation}

Where the kinematic chain $1$ represents the mobile base and $2$ the manipulator, thus, $q_1$ corresponds to the mobile base joints, i.e. the wheel driving $\theta_d$ and steering $\theta_s$ joints, and $q_2$ corresponds to the manipulator joints $\theta_m$, which are all rotational as aforementioned. 
As a result, the state transition matrix $A(n)$ is extracted straightforwardly from the generic one, but specifically for a platform with two kinematic chains. In particular, $I_1$, $V_1$ and $^w\mathcal{J}_1$ refer to the inertia, Coriolis/centrifugal and Jacobian matrices of the full-Ackermann non-holonomic mobile base, as well as $I_2$, $V_2$ and $^1\mathcal{J}_2$ refer to the 5DoF manipulator.
Remark that, as aforementioned, the mobile base Jacobian $^w\mathcal{J}_1$ is linerized by means of a TSL considering the notable non-linearity that appears due to the non-holonomic constraints.

The actuators of the system are the six driving and four steering joints of the wheels of the mobile base and the five rotational joints of the manipulator. As external disturbances, gravity acts on the mobile platform as a constant acceleration.
Thus, the control vector $u$ is defined as follows:

\begin{equation}
    u(n) = \left[
    \begin{gathered}
        \tau_d \:
        \tau_s \:
        \tau_m \:
        g \:
    \end{gathered}
    \right]^T
\end{equation}

Where $\tau_d$, $\tau_s$ and $\tau_m$ are the actuation torques to the driving, steering and manipulator joints respectively, and $g$ is the gravity acceleration.
Note that ExoTeR joints only receive position and velocity commands, nevertheless, using the whole dynamics model allows to generate torque-efficient motions.
Later on the joint position and speed commands are directly extracted from the state vector $x(n)$.

Once more, the input distribution matrix $B(n)$ is directly obtained using the generic one presented in section \ref{sec:algorithm} but with only one external disturbance, the gravity $g$.
On one hand, the effect of $g$ into the mobile base, $f^1_1$, generates a wheel-soil friction, which is modeled in a simplified way by means of the rolling resistance of the terrain as expressed in (\ref{eq:rolling_resistance}), with $\rho$ the rolling resistance coefficient of the terrain, $d_w$ the diameter of the wheels, $m$ the mass of the vehicle and $\mathcal{N}_w$ the number of wheels of the rover.
On the other hand, the effect of $g$ into the manipulator, $f^2_1$ is ignored, as aforementioned, considering the huge gear ratio of the arm joints.

\begin{equation}\label{eq:rolling_resistance}
    f^1_1 = \rho \: \frac{d_w}{2} \: \frac{m}{\mathcal{N}_w}
\end{equation}

Finally, several constraints have been defined to consider ExoTeR limits. On the one hand, the maximum actuation torque for the driving ($\tau_d$), steering ($\tau_s$) and manipulator ($\tau_m$) joints are included as state-input constraints. On the other hand, the limits on the velocity and acceleration of the driving ($\dot{\theta}_d$, $\ddot{\theta}_d$), the steering ($\dot{\theta}_s$, $\ddot{\theta}_s$) and the manipulator ($\dot{\theta}_m$, $\ddot{\theta}_m$) joints are defined as pure state constraints. Additionally, the position limits of the steering $\theta_s$ and manipulator $\theta_m$ joints are also included as pure state constraints.

\subsection{Replanning capability}

Given a feasible motion plan, i.e. the state $x(n)$ and control $u(n)$ vectors for the complete planning horizon $T$, a separate component needs to bring it to the mobile platform, ensuring it is properly followed until reaching the goal.
If there are deviations from what was planned, then this component has to take the right decisions to ensure that the goal is reached. 
This deviations can be caused by different means. The main ones are the model intrinsic errors because of the discretization and the linearization. But it is also pertinent to consider the effect of other agents into the system as external disturbances not considered initially in the model, like the platform localization error, the goal pose estimation error or the non ideal behaviour of the actuators.
In the particular SFR use case, the rover localization has a certain error, which adds up to the sample positioning error, induced by the sample detection and localization subsystem.
This sample positioning error is expected to be higher as the rover is far from it, arising the necessity of replanning the motion as the positioning error gets smaller, i.e. as the system gets closer to the sample.

Therefore, a motion plan follower has been developed, which implements a replanning capability similar to the event-triggered one proposed in \cite{Luis2020OnlinePlanning}, in the following manner. 
First, the motion plan follower sends sequentially the next actuation command to the platform (or the first one initially). 
Second, it checks if the goal pose has changed. If this is the case, then the system returns to the wave expansion of stage one (PPWS) and replans the whole motion. 
If not, a third step checks if there is too much drift in any of the controlled states, which would lead to returning to the trajectory extraction of stage one, which uses the current platform pose to replan the motion. 
Fourth, if no replan is needed and the goal is reached, the follower finishes the execution. 
Otherwise, it continues sending actuation commands in accordance to the already generated motion plan, and starts again the sequence.

\begin{figure}[t]
    \centering
    \includegraphics[width=\columnwidth]{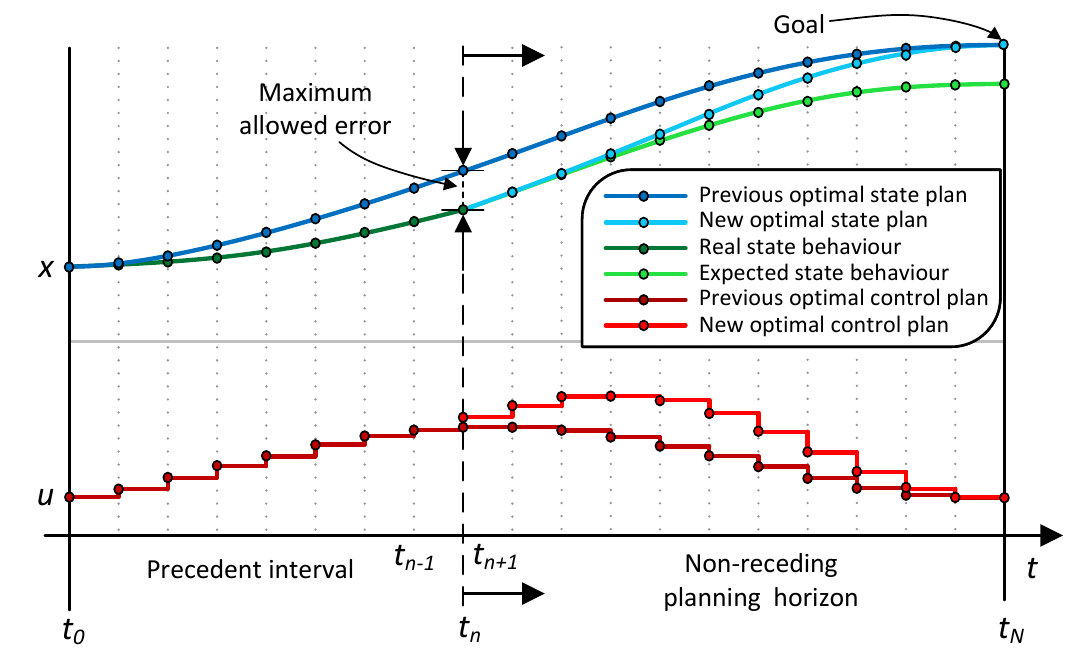}
    \caption{Graph exemplifying the replanning capability. The behaviour of a controlled state $x$ is continuously checked (dark green). If a considerable deviation from the previous plan (dark blue) is detected, a new global motion plan (light blue) is computed from the time step $t_n$ onwards. The new state and control plans (light red) allow the system to reach the goal smoothly correcting the previously accumulated error.}
    \label{fig:nonreceding}
\end{figure}

In case one of the states drifts from the planned motion, the behaviour of the replanning capability is exemplified in Figure \ref{fig:nonreceding}. 
Starting with a planned motion (dark blue) from $t_0$ to $t_N$, with $N$ number of time steps of $\Delta t$ size, and given the time evolution of a controlled state $x$ (dark green) in accordance to a given actuation $u$ (dark red), this evolution may differ from the plan, increasingly accumulating error.
When this error surpasses a certain threshold at time step $t_n$, the predicted behaviour of $x$ is completely undesired (light green), thus, a replan is launched using the previous motion plan, from $t_n$ onwards (dark blue), as a warm start. 
Thus, the replanned motion (light blue) compensates the accumulated drift in $x$ by slightly modifying the previous optimal control policy $u$ (dark red), generating a new one (light red) in the neighbourhood of the previous solution. In this way, the state $x$ will still reach the goal as long as the new motion plan is properly followed. 

\section{Results} \label{sec:results}

The proposed method for motion planning was validated by means of several tests with the SFR use case. 
On one hand, a deep performance analysis of the motion planner was performed with a benchmark between different layouts of the approach, i.e. using different combination of the already explained stages. 
This comparison confirmed that the complete motion planning approach is the most convenient, with a path planning warm start, a first unconstrained stage and a final constrained stage. 
On the other hand, several laboratory tests were performed with the Exomars Testing Rover (ExoTeR) in the Martian Analogue Testbed of the Planetary Robotics Laboratory (PRL) of the European Space Agency (ESA). 
Using MWMP and the proposed replanning procedure, ExoTeR was capable of successfully reach a martian sample tube and retrieve it with its manipulator.
The source code of the MWMP library used within these tests is available in MatLab\footnote{\url{https://github.com/spaceuma/MWMP-MatLab}} and C++\footnote{\url{https://github.com/spaceuma/MWMP-Cpp}}, under MIT open source license.

Thereupon, this section is divided into different subsections. 
First, the experimental setup, the scenario and the motion planner configuration are thoroughly detailed. 
Second, the performance benchmark is exposed. Third, the laboratory tests are presented.

\subsection{Experimental setup}
The goal of the performed laboratory tests was to demonstrate that ExoTeR can reach and retrieve a martian sample tube in a completely autonomous way.
For that purpose, these tests were carried out in the Martian Analogue Testbed at the ESA-PRL, as can be observed in Figure \ref{fig:exoter_depiction}. This is a 9x9 m experimental terrain which is highly representative of a real martian environment, including different types of soil (sandy, rocky), rocks or small slopes.

The tests should include real martian autonomous navigation restrictions in order to perform an illustrative retrieval of the SFR mission. 
Therefore, two additional subsystems were integrated in the platform, apart from the presented MWMP and replanning algorithms.
First, an autonomous sample detection and localization subsystem based on Convolutional Neural Networks (CNNs), which uses the LocCam stereo images to locate the sample tube with an average under $5\:cm$ position and $5^\circ$ orientation errors \cite{Castilla-Arquillo2022Hardware-acceleratedSimulations}. 
Second, a visual odometry algorithm for the platform localization, using the LocCam stereo camera and the IMU, with $7.5\:\%$ average localization drift in position and less than $2^\circ$ orientation error \cite{Geiger2011StereoScan:Real-time}. 
The Vicon markers were also used to obtain the ground-truth localization, not online but for data logging and post-processing purposes.

It was necessary to properly configure the SLQ motion planner for the tests.
The tests time horizon $t_N$ was $160\:s$, with a time step $\Delta t$ of $0.8\:s$. 
For the motion planner to converge, the maximum allowed position error was set-up to $1\:cm$ and the orientation error to $10^\circ$.
It was considered that the algorithm had converged if the norm of the stepped actuation $\bar{u}(n)$ was lower than $1\:\%$ of the norm of the whole actuation vector $u(n)$.
The particular costs which configured the LQR cost matrices $Q, R$ are defined in Table \ref{tab:lqr_costs}.
Note that the cost of modifying the input gravity disturbance $g$ in $R$ is the largest, to ensure that it remains as a constant gravity acceleration of $9.81 m/s^2$ precisely following the reference $u^0$.
Additionally, to ensure that the sample was retrieved perpendicularly to the ground, the goal pose orientation $^w\phi_{2}(N)$ was filled with roll $^w\varphi_{2}(N) = 0$, and the pitch $^w\vartheta_{2}(N)$ and yaw $^w\psi_{2}(N)$ were computed depending on the estimated sample orientation.

\begin{table}[t]
    \centering
    \caption{Quadratic costs configuration.}
    \vspace{1.5mm}
    {\renewcommand{\arraystretch}{1.5} 
    \begin{tabular}{| c | c | c |}
        \hline
        Type & Variable & Cost \\
        \hline
        \multirow{2}{*}{Goal state $Q(N)$} & EE pose $^wP_{2}$ & $10^{11}$ \\
            \cline{2-3}
            & Platform speed $^w\dot{P}_{1}$ & $10^6$\\
            \cline{2-3}
            & End effector speed $^1\dot{P}_{2}$ & $10^6$\\
        \hline
        \multirow{5}{*}{State full motion $Q(n)$} & Platform pose $^wP_{1}$ & $20$ \\
            \cline{2-3}
            & Driving wheels speed $\dot{\theta}_d$ & $100$\\
            \cline{2-3}
            & Driving wheels acceleration $\ddot{\theta}_d$ & $10^4$\\
            \cline{2-3}
            & Arm joints speed $\dot{\theta}_m$ & $3 \cdot 10^5$\\
            \cline{2-3}
            & Arm joints acceleration $\ddot{\theta}_m$ & $3 \cdot 10^5$\\
        \hline
        \multirow{4}{*}{Input full motion $R(n)$}
            & Wheels driving torque $\tau_{d}$ & $10^5$\\
            \cline{2-3}
            & Steering joints torque $\tau_{s}$ & $8\cdot10^4$\\
            \cline{2-3}
            & Arm joints torque $\tau_{m}$ & $10^{11}$ \\
            \cline{2-3}
            & Gravity $g$ & $10^{15}$ \\
        \hline
    \end{tabular} 
    }
    \label{tab:lqr_costs}
\end{table}


Lastly, the replanning was launched in accordance to certain errors when following the planned motions. 
In the first place, if the platform drifted more than $4\:cm$ from the planned path. 
In the second place, if any of the controlled joints deviated more than $2.29^\circ$ from the plan.
Additionally, every time the goal sample pose differed more than $3\:cm$ or $17.19^\circ$ from the previous estimation, a complete motion replan was launched.

\subsection{Motion planner performance analysis}
The performance of the proposed motion planning approach was analyzed using the aforementioned use case and setup, to showcase its advantages w.r.t. any possible layout of the stages, for instance, cold started or single-staged versions of the motion planner. 
For that purpose, six different layouts of the motion planner were defined, which include every possible combination of the three stages (USLQ, PPWS+USLQ, CSLQ, PPWS+CSLQ, USLQ+CSLQ) and the complete approach (PPWS+USLQ+CSLQ = MWMP).
For every layout, the same 21 motion plans were launched, using the PRL scenario with the ExoTeR model and the SFR use case, changing the initial rover pose and the goal sample pose. 

\begin{figure}[t]
    \centering
    \includegraphics[width=\columnwidth]{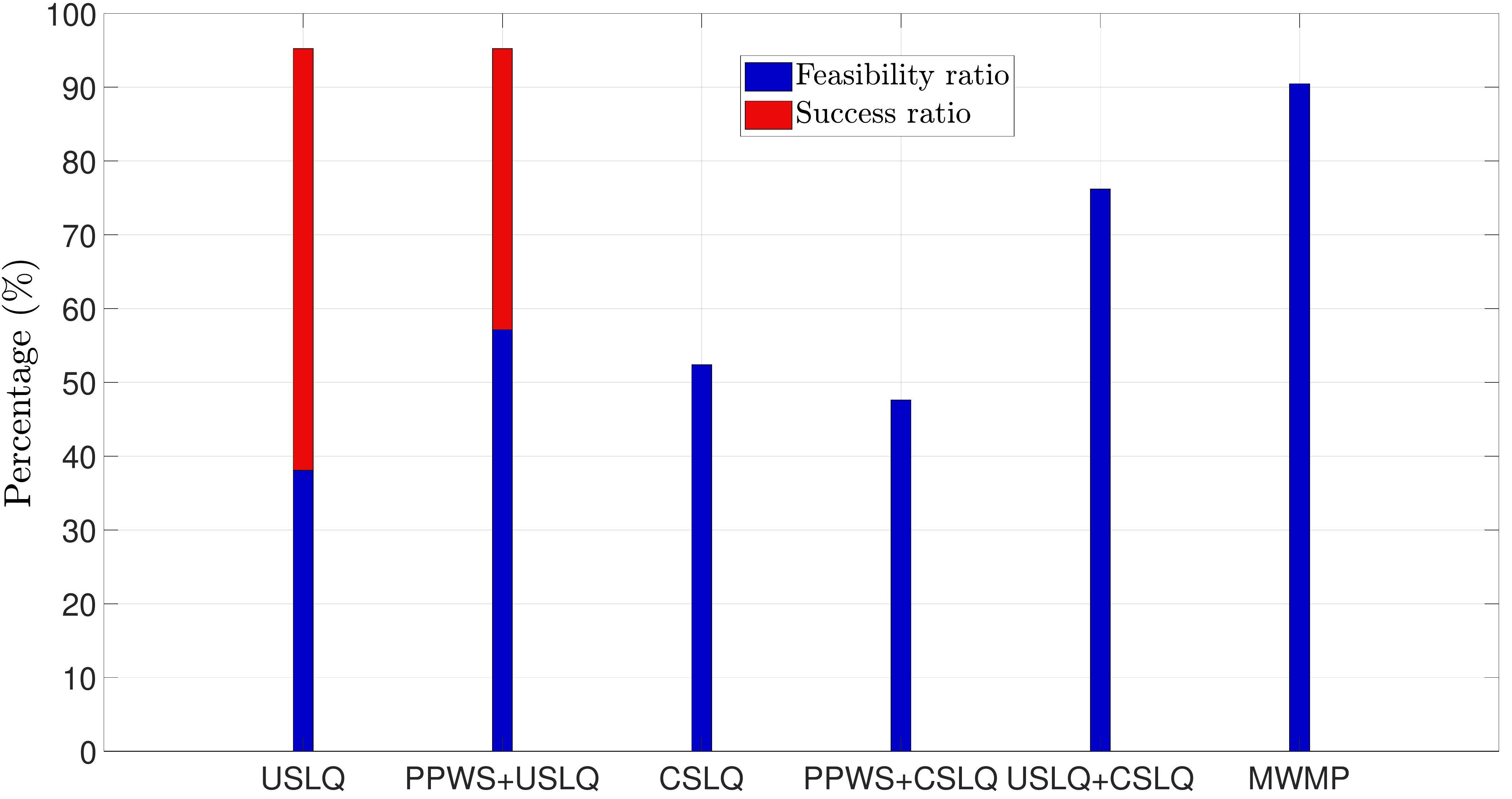}
    \caption{Measured success and feasibility ratios on the performance tests. For every layout, 21 different motion plans were launched. Note that the layouts that include the constrained stage (CSLQ) have the same percentage of successful and feasible motion plans, since this stage ensures feasibility if the algorithm converges.}
    \label{fig:feasibility}
\end{figure}

\begin{figure}[t]
    \centering
    \includegraphics[width=\columnwidth]{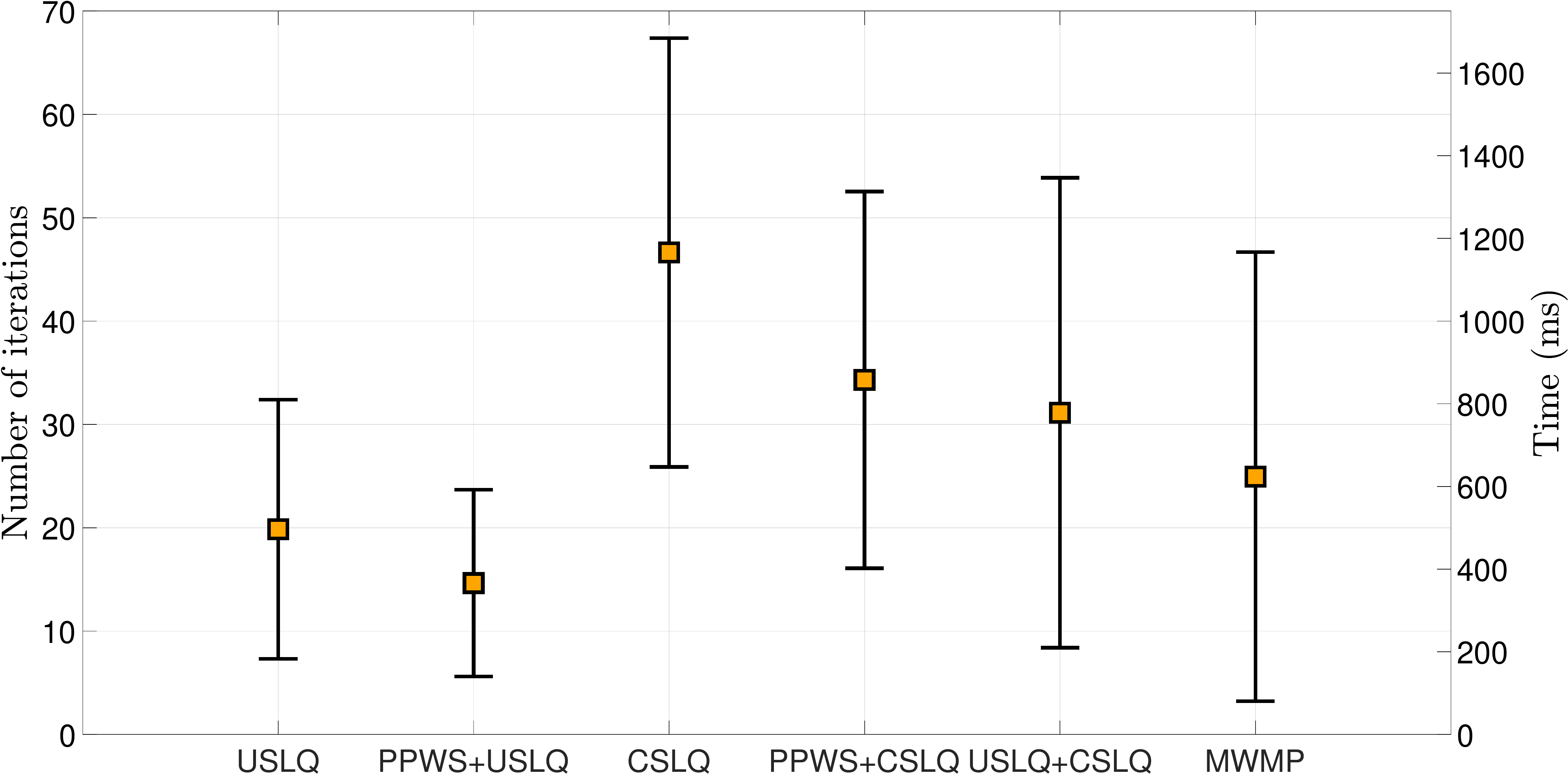}
    \caption{Measured iterations and execution time on the performance tests. For every layout, 21 different motion plans were launched. Average measurements are shown, including the standard deviation on the samples.}
    \label{fig:iterations}
\end{figure}

Three main parameters were measured within the tests.
First, the success rate, as a percentage. This represents the ratio of successful motion plans in the 21 tests, i.e. when the algorithm converges in less than 100 iterations, which is shown in Figure \ref{fig:feasibility}.
Second, the feasibility rate, which represents the percentage of constraint compliant motion plans in the 21 tests, which is also shown in Figure \ref{fig:feasibility}.
Clearly, the feasibility rate encompasses the success rate, given that a motion plan can only be feasible if the solver converges, i.e. if the motion plan is successful. Hence, every layout that makes use of the CSLQ stage has equal success and feasibility rates, since the algorithm only converges if the constraints are fulfilled.
Third, the average number of iterations until convergence. If several stages are established, then the total number of iterations is used. 
The extracted results regarding the number of iterations are shown in Figure \ref{fig:iterations}.
Note that the number of iterations is employed in the following as a measure of the convergence speed, since the computational time spent is not representative due to its dependency on external factors, such as the hardware, the quality of the software implementation or the CPU usage. 
Besides, the computational time spent is nearly proportional to the number of iterations, as an example, spending approximately $25ms$ per iteration in these tests, run on a single core of an Intel(R) Core(TM) i7-10750H CPU (2.60GHz).

\subsection{Lab tests}
To validate the proposed motion planner approach and the feasibility of the generated motion plans, several analogue SFR experiments were carried out at the Martian Analogue Testbed at ESA-PRL, starting from different rover locations and with different sample poses.
Four of the most representative tests are analyzed in this paper, and one of them is shown in a summarizing video\footnote{\url{https://youtu.be/xDFv4Ho4KZs}} of the lab tests campaign.
Additionally, an example of the evolution of the tests is also shown in Figure \ref{fig:motions_comparison}.

The experiments were run as follows. First, it was assumed that the sample was inside ExoTeR's LocCam Field of View (FoV). Therefore, the sample detection and localization subsystem was launched at the beginning to provide the initial estimation of the sample pose. 
Then, the sample pose was translated into an end effector goal pose, just above the sample and approaching the ground perpendicularly, and this goal pose was fed to MWMP to compute a global initial motion plan.
This was sent to the motion plan follower, which started to send the control commands at each time step, and receive the robot measured state.
As explained in Section \ref{sec:use_case}, the follower continuously checked if any of the controlled states was accumulating too much drift, given the defined thresholds.
Then, if necessary, the motion was replanned using the last motion plan as a warm start to accelerate the computation.
Additionally, the sample detection and localization subsystem was launched repeatedly with a frequency of $0.1\:Hz$, in order to keep improving the sample pose estimation and filtering the localization error due to the use of visual odometry, eventually triggering additional motion replans. 
Once the end effector had reached its goal pose, the execution is finished, and a separate sample retrieval component was launched just to perform the final sample grasping movement.

\section{Discussion}\label{sec:discussion}

\begin{table*}[t]
    \centering
    \caption{Lab test results}
    \vspace{1mm}
    \begin{tabular}{| c | c | c | c | c |}
        \hline
        SFR Test Case & 1 & 3 & 4 & 5 \\
        \hline
        First plan number of iterations & 17 
                                        & 20 
                                        & 29 
                                        & 13 \\
        \hline
        Number of replans (after goal changed) & 6 (6) 
                          & 5 (3)
                          & 7 (2)
                          & 7 (5) \\
        \hline
        Average number of iterations & 13.43 
                                     & 16.60 
                                     & 25.62
                                     & 60.33 \\
        \hline
        Average arm joints position error ($^\circ$) & 0.2521 
                                                & 0.2235 
                                                & 0.9167
                                                & 0.1719 \\
        \hline
        Average steering joints position error ($^\circ$) & 2.6471 
                                                     & 0.0521 
                                                     & 0.0997 
                                                     & 0.0555 \\
        \hline
        Rover base final pose error (m, $^\circ$) & 0.0023, 2.0798 
                                             & 0.036, 0.4183 
                                             & 0.02, 0.0
                                             & 0.0361, 0.0630 \\
        \hline
        End effector final pose error (m, $^\circ$) & 0.0942, 15.8996 
                                              & 0.0221, 12.4332 
                                              & 0.0441, 0.1833
                                              & 0.0332, 24.9007 \\
        \hline
        Sample pose estimation error (m, $^\circ$) & 0.0908, 15.7964 
                                              & 0.0150, 12.0035 
                                              & 0.0220, 0.6303
                                              & 0.0030, 24.8893 \\
        \hline

    \end{tabular} 
    \label{tab:labtest_results}
\end{table*}

On the one hand, concerning the performance tests, their results are summarized in Figures \ref{fig:feasibility} and \ref{fig:iterations}. 
As can be observed, the path planner warm start (PPWS+USLQ, PPWS+CSLQ, PPWS+USLQ+CSLQ) always reduces the average number of iterations w.r.t. the cold started versions (USLQ, CSLQ, USLQ+CSLQ), reducing also the convergence speed variability. This means that the motion planner behaviour is more predictable.
Additionally, the fastest layouts are the unconstrained ones (USLQ, $19.85$ it; PPWS+USLQ, $14.65$ it), as expected and aforementioned. Although these layouts have really high success ratios (USLQ, $95.24\%$; PPWS+USLQ, $95.24\%$), they can not guarantee constraints compliance, thus they also have the lowest feasibility ratios (USLQ, $38.10\%$; PPWS+USLQ, $57.14\%$).
The slowest layouts are the constrained ones (CSLQ, $46.645$ it; PPWS+CSLQ, $34.31$ it), and they still do not reach high feasibility ratios (CSLQ, $52.38\%$; PPWS+CSLQ, $47.62\%$), due to their convergence difficulties for the highly over-actuated and constrained SFR use case.
Finally, the unconstrained-constrained layouts have an intermediate convergence speed (USLQ+CSLQ, $31.13$ it; MWMP, $24.95$ it), having the complete approach (MWMP) a comparable convergence speed to the unconstrained layouts.
Besides, these layouts also have the highest feasibility ratios (USLQ+CSLQ, $76.19\%$; MWMP, $90.48\%$), thanks to the successive warm start procedure.

Summarizing, the performance tests demonstrate that the proposed multi-staged approach (MWMP) improves the behaviour of the optimal motion planner, increasing noticeably the average number of feasible motion plans ($90.48\%$) maintaining a considerably low average number of iterations until convergence ($24.95$ it).

On the other hand, the results of four of the lab tests are summarized in Table \ref{tab:labtest_results}. 
First, the initial motion plan number of iterations (20 it. avg) matches what it is expected, seeing the results of the performance tests (25 it. avg). 
Furthermore, the average number of iterations including the replans is generally lower than the first plan iterations, which confirms that using the last motion plan as a warm start accelerates the motion planning procedure.
This did not happen in case 5, since a sharp turn was required and the steering joints were repeatedly reaching their limits.
It is remarkable that the average errors w.r.t. the planned motion of the controlled states, i.e. the arm joints ($0.3896 ^\circ$ avg) and the steering joints ($0.7105 ^\circ$ avg), are negligible, which means that the predicted motion was accurate and the discretization errors do not severely impact the system behaviour, thanks to a sufficiently small time step $\Delta t$.
Besides, most of the required replans were performed due to the low accuracy of the sample pose estimator (16 out of 25), which changed the goal pose substantially several times during the tests, as it can be observed in Table \ref{tab:labtest_results}.
Regarding the non directly controlled states, the errors of the rover base ($0.0236 m$, $0.6417 ^\circ$ avg) and the end effector ($0.0484 m$, $13.3556 ^\circ$ avg) final poses are also small, therefore, it is confirmed that the system model is representative, and the motion planner is accurate enough to ensure a successful sample retrieval, considering the $7 cm$ full opened gripper width.
Note also that the end effector final pose error is caused mainly by the sample pose estimator, being minimal the errors induced by the motion planner itself.

\begin{figure*}[t]
    \subfloat[]{
        \includegraphics[width=0.24\textwidth]{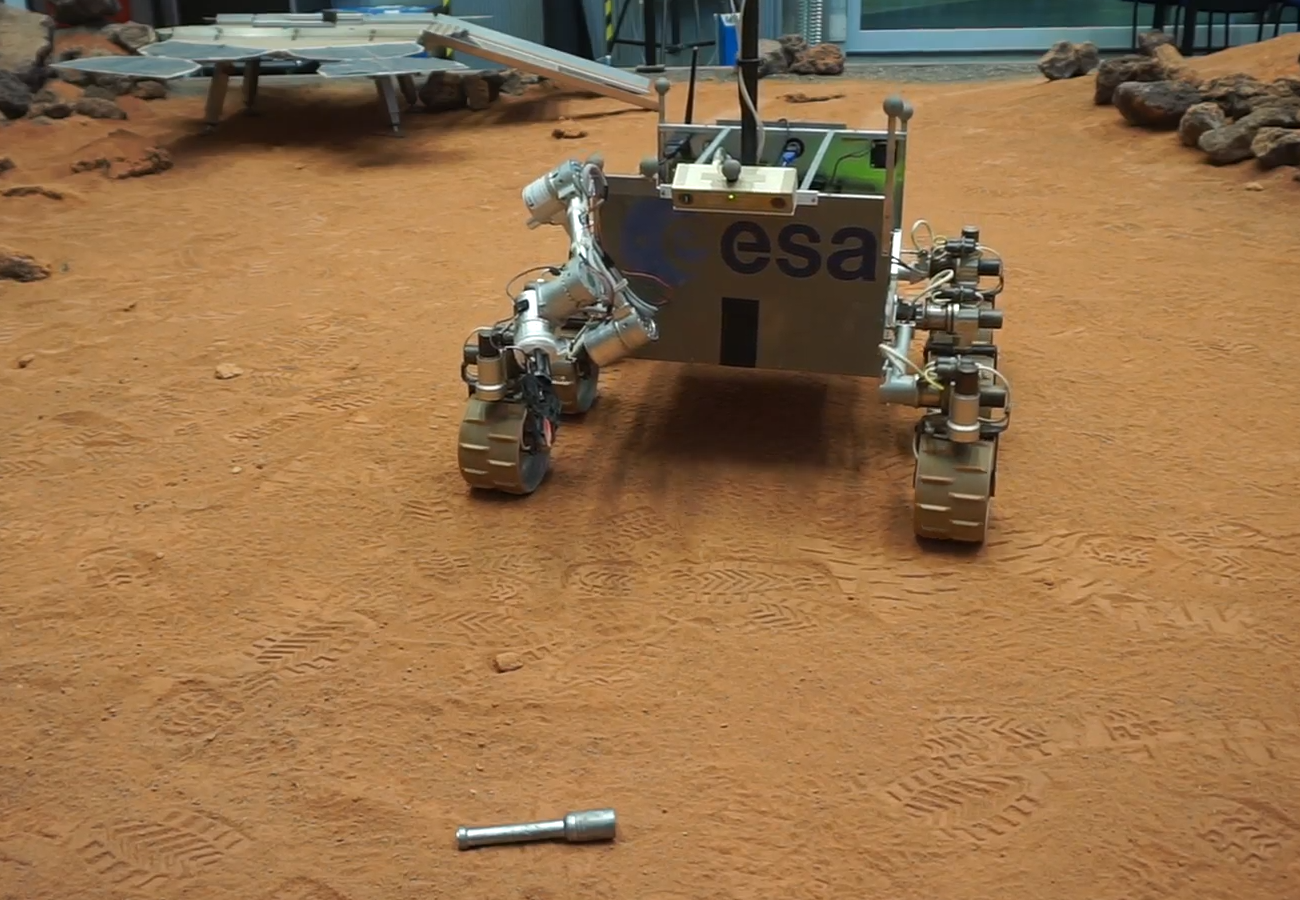}
        \label{fig:standard00_approaching}
    }
    \subfloat[]{
        \includegraphics[width=0.24\textwidth]{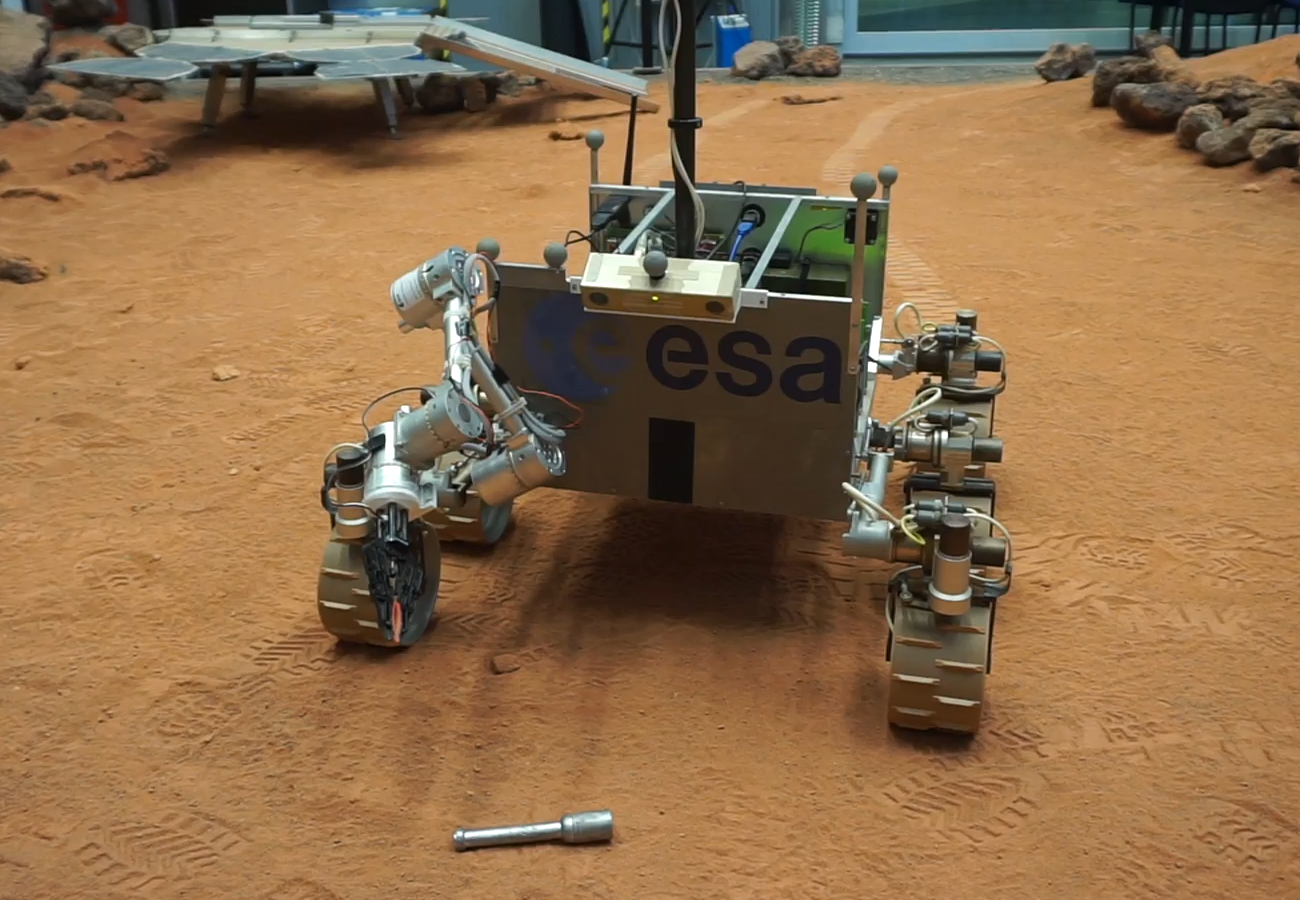}
        \label{fig:standard01_reached}
    }
    \subfloat[]{
        \includegraphics[width=0.24\textwidth]{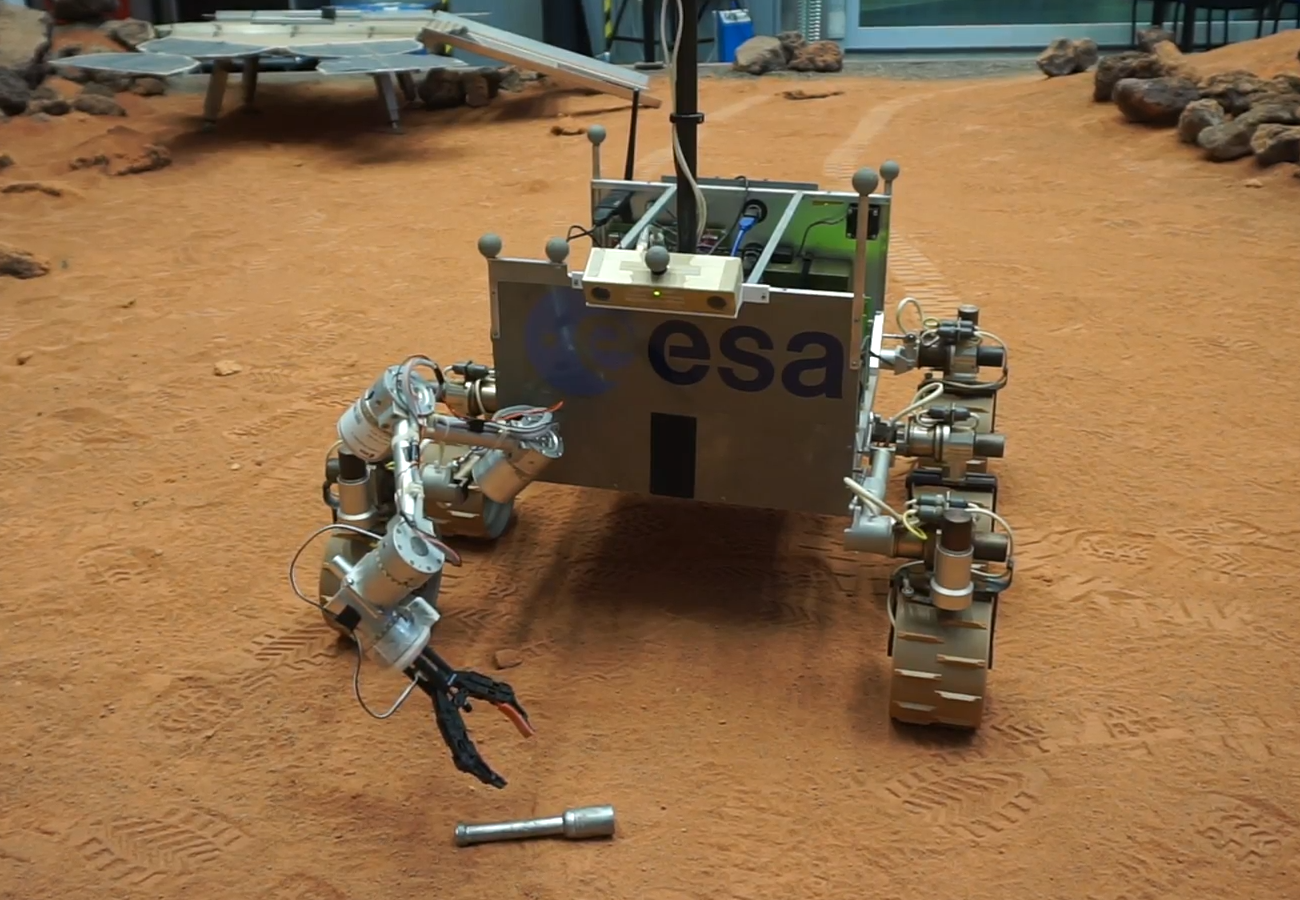}
        \label{fig:standard02_above}
    }
    \subfloat[]{
        \includegraphics[width=0.24\textwidth]{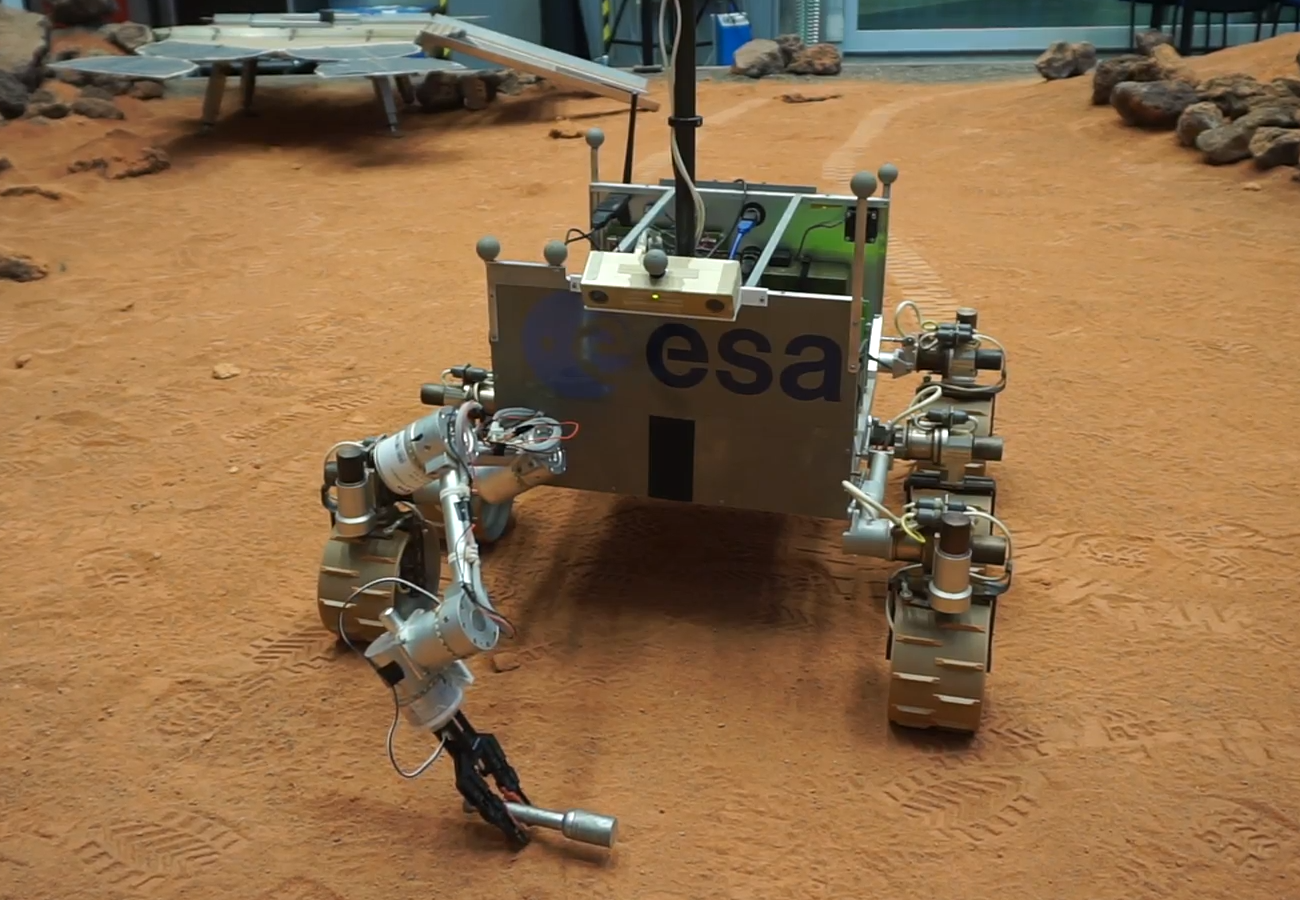}
        \label{fig:standard04_gripped}
    }
    \\
    \subfloat[]{
        \includegraphics[width=0.24\textwidth]{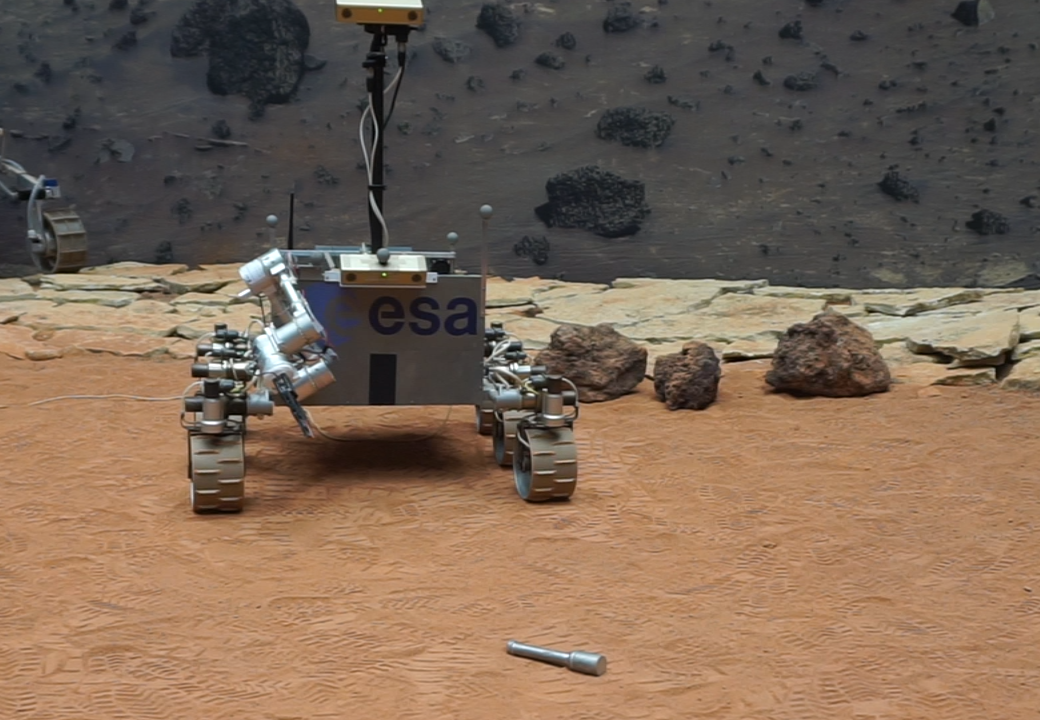}
        \label{fig:optimal00_approaching}
    }
    \subfloat[]{
        \includegraphics[width=0.24\textwidth]{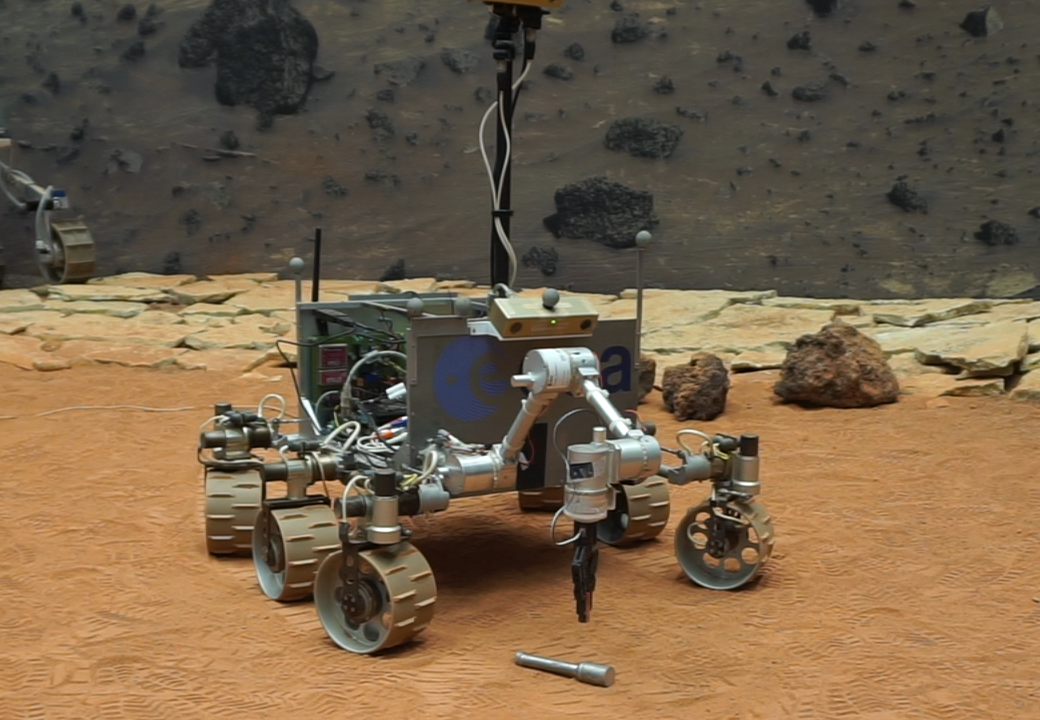}
        \label{fig:optimal01_reached}
    }
    \subfloat[]{
        \includegraphics[width=0.24\textwidth]{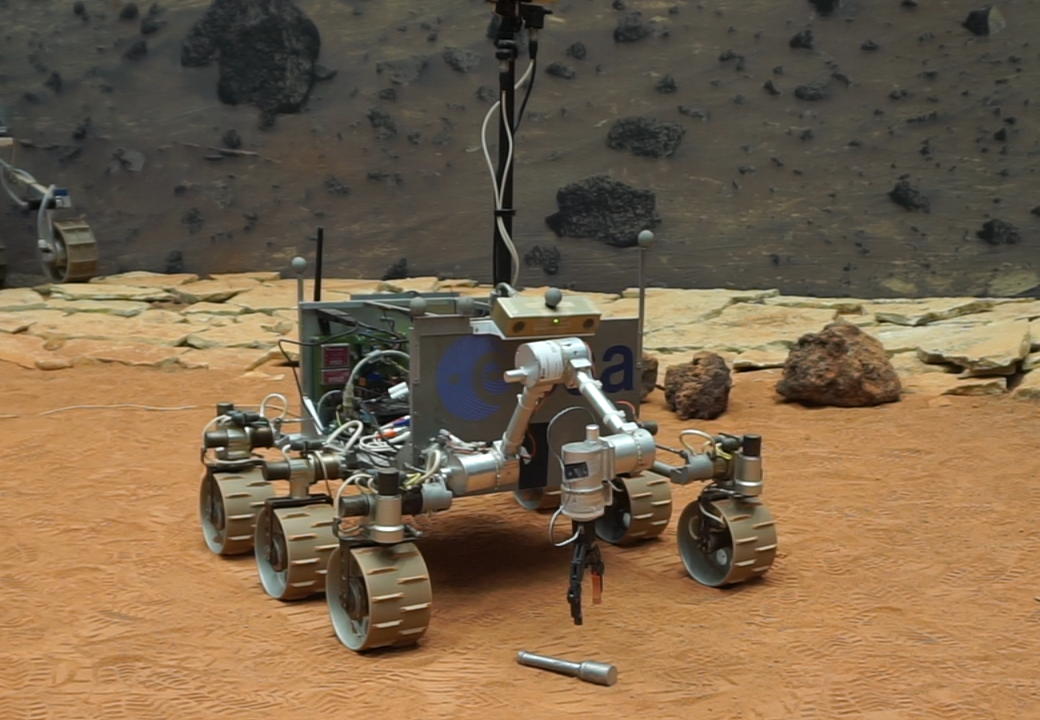}
        \label{fig:optimal02_above}
    }
    \subfloat[]{
        \includegraphics[width=0.24\textwidth]{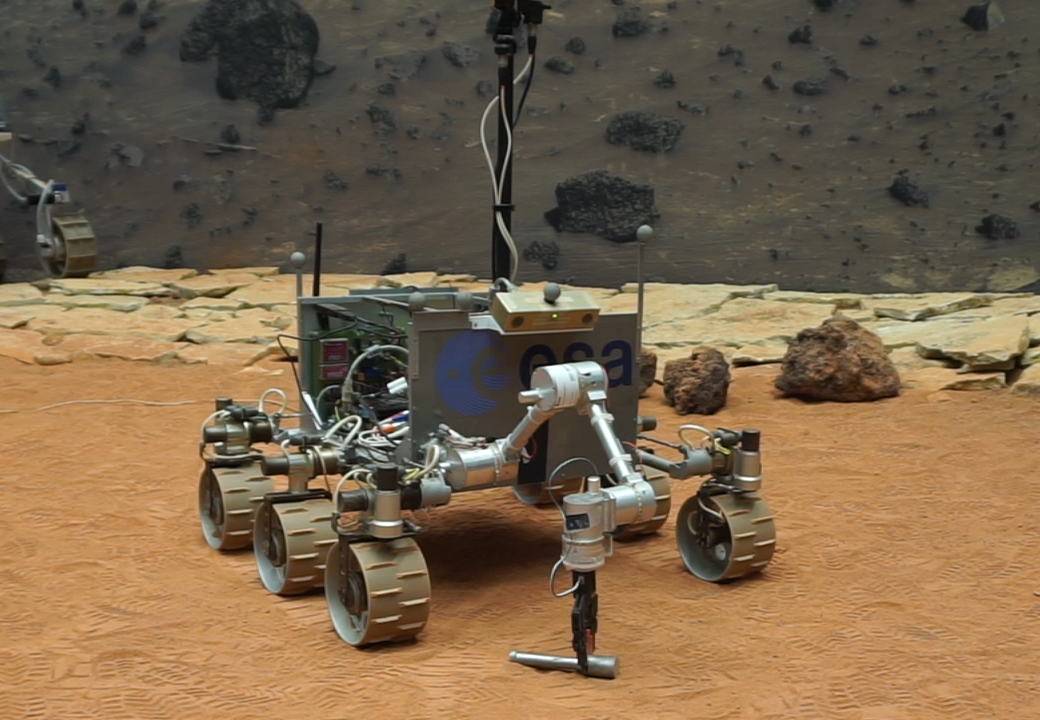}
        \label{fig:optimal04_gripped}
    }
    \caption{Motion evolution of ExoTeR during two different SFR tests, using a decoupled motion planning approach (a-d) or MWMP (e-h). The decoupled solution is not prepared to retrieve the sample once it reaches it (b), and can not retrieve the sample perpendicularly to the ground (c), with the result of a defective grasp (d). On the other hand, MWMP generates an optimal motion, leaving the arm prepared to the retrieval operation as soon as the base stops (f), being it placed in a certain pose which allows the manipulator to retrieve the sample perpendicularly (g), with a high quality grasp (h).}
    \label{fig:motions_comparison}
\end{figure*}

Finally, the evolution of the SFR Test Case 3 is shown in Figures \ref{fig:optimal00_approaching}-\ref{fig:optimal04_gripped}, in comparison to another SFR test in Figures \ref{fig:standard00_approaching}-\ref{fig:standard04_gripped}, performed with a standard non-optimal and decoupled motion planning approach \cite{Mantoani2022SamplesRover}.
As can be observed, although both approaches start from similar situations (Figures \ref{fig:standard00_approaching} and \ref{fig:optimal00_approaching}), MWMP leaves the manipulator prepared for the retrieval operation as soon as the rover reaches the sample (Figure \ref{fig:optimal01_reached}), meanwhile the decoupled solution yet requires to move the arm once the rover stops (Figure \ref{fig:standard01_reached}).
Additionally, the decoupled solution does not place the rover base in a good position considering the posterior retrieval operation, thus, the gripper orientation is not perpendicular to the ground and does not match the sample yaw (Figure \ref{fig:standard02_above}), which, in the end, generates a defective grasp (Figure \ref{fig:standard04_gripped}).
Conversely, the optimal motion planner uses all the system joints (rover and manipulator), placing the base to leave the arm in a perfectly perpendicular pose w.r.t. the sample. 
Therefore, the gripper is orientated perpendicularly to the ground and matching the sample yaw (Figure \ref{fig:optimal02_above}), being the quality of the grasp, thus, much higher (Figure \ref{fig:optimal04_gripped}).

\section{Conclusion} \label{sec:conclusions}

In this paper MWMP is presented, a motion planner for over-actuated mobile platforms capable of dealing with non-linearities, such as non-holonomic constraints or joints limits, without severely impacting the computational resources of the system.
This is achieved by means of a multi-staged warm start approach, which initializes in several steps the optimal solver, SLQ.
In particular, three different stages are used: first, a FMM based path planner; second, an unconstrained SLQ motion planner; third, a constrained SLQ motion planner.
This complete approach has been demonstrated to improve the performance of the motion planner in comparison with any other combination of the stages, since the algorithm converges faster and the probability of finding a feasible solution is the highest (up to twice as fast and $40\%$ more feasible solutions in comparison with the standard Constrained SLQ).

Furthermore, a generic state space model for over-actuated mobile platforms has been presented, which can model platforms composed of several kinematic chains in a straightforward way.
This model is particularized for the rover ExoTeR, composed by a mobile base and a robotic arm, which is later used to perform some laboratory tests to validate the motion planner.
For that purpose, a tailored event-triggered replanning capability has been included, which allows the system to precisely follow the generated motion plans.
The performed laboratory tests emulate a SFR-like sample retrieval mission, showcasing the advantages of the presented motion planner to generate accurate motions for highly redundant and non-linear platforms, in this case, allowing ExoTeR to retrieve a Mars soil sample tube with a high quality grasp, even considering the rover and sample localization errors.

MWMP has demonstrated its capacity to deal with high kinematic redundancy, using the rover base joints to improve the quality of the later manipulator grasping operation. Nevertheless, it is foreseen that MWMP could handle joint breakdowns, which will be tested in the hereafter.
Lastly, it is expected the use of this motion planner in further use cases, including 3D platforms with faster dynamics and multiple external disturbances. Thus, a benchmark of the proposed replanning capability with other methodologies, such as existing MPC controllers, is also planned as future work.




\bibliographystyle{IEEEtran.bst}
\bibliography{references.bib}

\begin{thebibliography}{10}
\providecommand{\url}[1]{#1}
\csname url@samestyle\endcsname
\providecommand{\newblock}{\relax}
\providecommand{\bibinfo}[2]{#2}
\providecommand{\BIBentrySTDinterwordspacing}{\spaceskip=0pt\relax}
\providecommand{\BIBentryALTinterwordstretchfactor}{4}
\providecommand{\BIBentryALTinterwordspacing}{\spaceskip=\fontdimen2\font plus
\BIBentryALTinterwordstretchfactor\fontdimen3\font minus
  \fontdimen4\font\relax}
\providecommand{\BIBforeignlanguage}[2]{{%
\expandafter\ifx\csname l@#1\endcsname\relax
\typeout{** WARNING: IEEEtran.bst: No hyphenation pattern has been}%
\typeout{** loaded for the language `#1'. Using the pattern for}%
\typeout{** the default language instead.}%
\else
\language=\csname l@#1\endcsname
\fi
#2}}
\providecommand{\BIBdecl}{\relax}
\BIBdecl

\bibitem{Latombe1991RobotPlanning}
J.-C. Latombe, \emph{{Robot Motion Planning}}.\hskip 1em plus 0.5em minus
  0.4em\relax Boston, MA: Springer US, 1991.

\bibitem{Alatise2020AMethods}
M.~B. Alatise and G.~P. Hancke, ``{A Review on Challenges of Autonomous Mobile
  Robot and Sensor Fusion Methods},'' \emph{IEEE Access}, vol.~8, pp.
  39\,830--39\,846, 2020.

\bibitem{Soni2018FormationSurvey}
A.~Soni and H.~Hu, ``{Formation Control for a Fleet of Autonomous Ground
  Vehicles: A Survey},'' \emph{Robotics 2018, Vol. 7, Page 67}, vol.~7, no.~4,
  p.~67, 11 2018.

\bibitem{Gerdes2020EfficientResources}
L.~Gerdes, M.~Azkarate, J.~R. S{\'{a}}nchez-Ib{\'{a}}{\~{n}}ez, L.~Joudrier,
  and C.~J. Perez-del Pulgar, ``{Efficient autonomous navigation for planetary
  rovers with limited resources},'' \emph{Journal of Field Robotics}, vol.~37,
  no.~7, pp. 1153--1170, 10 2020.

\bibitem{Wang2019Roboat:Waterways}
W.~Wang, B.~Gheneti, L.~A. Mateos, F.~Duarte, C.~Ratti, and D.~Rus, ``{Roboat:
  An Autonomous Surface Vehicle for Urban Waterways},'' \emph{IEEE
  International Conference on Intelligent Robots and Systems}, pp. 6340--6347,
  11 2019.

\bibitem{Kratky2021AutonomousVehicles}
V.~Kratky, A.~Alcantara, J.~Capitan, P.~Stepan, M.~Saska, and A.~Ollero,
  ``{Autonomous Aerial Filming with Distributed Lighting by a Team of Unmanned
  Aerial Vehicles},'' \emph{IEEE Robotics and Automation Letters}, vol.~6,
  no.~4, pp. 7580--7587, 10 2021.

\bibitem{Huang2018EfficientVehicle}
S.~W. Huang, E.~Chen, and J.~Guo, ``{Efficient Seafloor Classification and
  Submarine Cable Route Design Using an Autonomous Underwater Vehicle},''
  \emph{IEEE Journal of Oceanic Engineering}, vol.~43, no.~1, pp. 7--18, 1
  2018.

\bibitem{Araguz2018ApplyingProspects}
C.~Araguz, E.~Bou-Balust, and E.~Alarc{\'{o}}n, ``{Applying autonomy to
  distributed satellite systems: Trends, challenges, and future prospects},''
  \emph{Systems Engineering}, vol.~21, no.~5, pp. 401--416, 9 2018.

\bibitem{Pierre2019TowardRobots}
R.~St.~Pierre and S.~Bergbreiter, ``Toward autonomy in sub-gram terrestrial
  robots,'' \emph{Annual Review of Control, Robotics, and Autonomous Systems},
  vol.~2, pp. 231--252, 2019.

\bibitem{Pilania2018MobileEnvironments}
V.~Pilania and K.~Gupta, ``{Mobile manipulator planning under uncertainty in
  unknown environments},'' \emph{The International Journal of Robotics
  Research}, vol.~37, no. 2-3, pp. 316--339, 2 2018.

\bibitem{Pajak2017Point-to-PointManipulators}
G.~Pajak and I.~Pajak, ``Point-to-point collision-free trajectory planning for
  mobile manipulators,'' \emph{Journal of Intelligent \& Robotic Systems},
  vol.~85, no.~3, pp. 523--538, 2017.

\bibitem{Li2020AManipulator}
Q.~Li, Y.~Mu, Y.~You, Z.~Zhang, and C.~Feng, ``{A Hierarchical Motion Planning
  for Mobile Manipulator},'' \emph{IEEJ Transactions on Electrical and
  Electronic Engineering}, vol.~15, no.~9, pp. 1390--1399, 9 2020.

\bibitem{Liao2019Optimization-basedRedundancy}
J.~Liao, F.~Huang, Z.~Chen, and B.~Yao, ``{Optimization-based motion planning
  of mobile manipulator with high degree of kinematic redundancy},''
  \emph{International Journal of Intelligent Robotics and Applications},
  vol.~3, no.~2, pp. 115--130, 6 2019.

\bibitem{Chen2019}
J.~Chen, W.~Zhan, and M.~Tomizuka, ``{Autonomous driving motion planning with
  constrained iterative LQR},'' \emph{IEEE Transactions on Intelligent
  Vehicles}, vol.~4, no.~2, pp. 244--254, 6 2019.

\bibitem{Preda2021OptimalProgramming}
V.~Preda, A.~Hyslop, and S.~Bennani, ``{Optimal science-time reorientation
  policy for the Comet Interceptor flyby via sequential convex programming},''
  \emph{CEAS Space Journal 2021}, vol.~1, pp. 1--14, 6 2021.

\bibitem{Foehn2021Time-optimalFlight}
P.~Foehn, A.~Romero, and D.~Scaramuzza, ``{Time-optimal planning for quadrotor
  waypoint flight},'' \emph{Science Robotics}, vol.~6, no.~56, 7 2021.

\bibitem{Minniti2019}
M.~V. Minniti, F.~Farshidian, R.~Grandia, and M.~Hutter, ``{Whole-Body MPC for
  a Dynamically Stable Mobile Manipulator},'' \emph{IEEE Robotics and
  Automation Letters}, vol.~4, no.~4, pp. 3687--3694, 10 2019.

\bibitem{Giftthaler2017}
M.~Giftthaler and J.~Buchli, ``{A projection approach to equality constrained
  iterative linear quadratic optimal control},'' in \emph{IEEE-RAS
  International Conference on Humanoid Robots}.\hskip 1em plus 0.5em minus
  0.4em\relax IEEE Computer Society, 12 2017, pp. 61--66.

\bibitem{Farshidian2017}
F.~Farshidian, E.~Jelavic, A.~Satapathy, M.~Giftthaler, and J.~Buchli,
  ``{Real-Time motion planning of legged robots: A model predictive control
  approach},'' in \emph{IEEE-RAS International Conference on Humanoid
  Robots}.\hskip 1em plus 0.5em minus 0.4em\relax IEEE Computer Society, 12
  2017, pp. 577--584.

\bibitem{Fnadi2021ConstrainedGrounds}
M.~Fnadi, W.~Du, F.~Plumet, and F.~Benamar, ``{Constrained Model Predictive
  Control for dynamic path tracking of a bi-steerable rover on slippery
  grounds},'' \emph{Control Engineering Practice}, vol. 107, p. 104693, 2 2021.

\bibitem{Neunert2016}
M.~Neunert, C.~De~Crousaz, F.~Furrer, M.~Kamel, F.~Farshidian, R.~Siegwart, and
  J.~Buchli, ``Fast nonlinear model predictive control for unified trajectory
  optimization and tracking,'' in \emph{2016 IEEE international conference on
  robotics and automation (ICRA)}.\hskip 1em plus 0.5em minus 0.4em\relax IEEE,
  2016, pp. 1398--1404.

\bibitem{Wang2021TransportingFramework}
Y.~Wang, H.~Kusano, and T.~Sugihara, ``{Transporting a heavy object on a
  frictional floor by a mobile manipulator based on adaptive MPC framework},''
  \emph{2021 IEEE/SICE International Symposium on System Integration, SII
  2021}, pp. 807--812, 1 2021.

\bibitem{Colombo2019ParameterizedApproach}
R.~Colombo, F.~Gennari, V.~Annem, P.~Rajendran, S.~Thakar, L.~Bascetta, and
  S.~K. Gupta, ``{Parameterized model predictive control of a nonholonomic
  mobile manipulator: A terminal constraint-free approach},'' \emph{IEEE
  International Conference on Automation Science and Engineering}, vol.
  2019-August, pp. 1437--1442, 8 2019.

\bibitem{Sleiman2021AManipulation}
J.~P. Sleiman, F.~Farshidian, M.~V. Minniti, and M.~Hutter, ``{A Unified MPC
  Framework for Whole-Body Dynamic Locomotion and Manipulation},'' \emph{IEEE
  Robotics and Automation Letters}, vol.~6, no.~3, pp. 4688--4695, 7 2021.

\bibitem{Younes2021NonlinearGeneration}
Y.~A. Younes and M.~Barczyk, ``{Nonlinear Model Predictive Horizon for Optimal
  Trajectory Generation},'' \emph{Robotics 2021, Vol. 10, Page 90}, vol.~10,
  no.~3, p.~90, 7 2021.

\bibitem{Luis2020OnlinePlanning}
C.~E. Luis, M.~Vukosavljev, and A.~P. Schoellig, ``{Online Trajectory
  Generation with Distributed Model Predictive Control for Multi-Robot Motion
  Planning},'' \emph{IEEE Robotics and Automation Letters}, vol.~5, no.~2, pp.
  604--611, 4 2020.

\bibitem{Plancher2021AcceleratingFPGA}
B.~Plancher, S.~M. Neuman, T.~Bourgeat, S.~Kuindersma, S.~Devadas, and V.~J.
  Reddi, ``{Accelerating Robot Dynamics Gradients on a CPU, GPU, and FPGA},''
  \emph{IEEE Robotics and Automation Letters}, vol.~6, no.~2, pp. 2335--2342, 4
  2021.

\bibitem{Bitar2019Warm-StartedASVs}
G.~Bitar, V.~N. Vestad, A.~M. Lekkas, and M.~Breivik, ``{Warm-Started Optimized
  Trajectory Planning for ASVs},'' \emph{IFAC-PapersOnLine}, vol.~52, no.~21,
  pp. 308--314, 1 2019.

\bibitem{Zhang2019AutonomousAvoidance}
X.~Zhang, A.~Liniger, A.~Sakai, and F.~Borrelli, ``{Autonomous Parking Using
  Optimization-Based Collision Avoidance},'' \emph{Proceedings of the IEEE
  Conference on Decision and Control}, vol. 2018-December, pp. 4327--4332, 1
  2019.

\bibitem{Lembono2020MemoryOptimization}
T.~S. Lembono, A.~Paolillo, E.~Pignat, and S.~Calinon, ``{Memory of Motion for
  Warm-Starting Trajectory Optimization},'' \emph{IEEE Robotics and Automation
  Letters}, vol.~5, no.~2, pp. 2594--2601, 4 2020.

\bibitem{Mansard2018UsingController}
N.~Mansard, A.~Delprete, M.~Geisert, S.~Tonneau, and O.~Stasse, ``{Using a
  Memory of Motion to Efficiently Warm-Start a Nonlinear Predictive
  Controller},'' \emph{Proceedings - IEEE International Conference on Robotics
  and Automation}, pp. 2986--2993, 9 2018.

\bibitem{Ichnowski2020DeepPlanning}
J.~Ichnowski, Y.~Avigal, V.~Satish, and K.~Goldberg, ``{Deep learning can
  accelerate grasp-optimized motion planning},'' \emph{Science Robotics},
  vol.~5, no.~48, 11 2020.

\bibitem{Sethian1999}
J.~A. Sethian, ``{Fast Marching Methods},'' \emph{SIAM Review}, vol.~41, no.~2,
  pp. 199--235, 1999.

\bibitem{Sideris2005AnProblems}
A.~Sideris and J.~E. Bobrow, ``{An efficient sequential linear quadratic
  algorithm for solving nonlinear optimal control problems},'' \emph{IEEE
  Transactions on Automatic Control}, vol.~50, no.~12, pp. 2043--2047, 12 2005.

\bibitem{Sideris2011}
A.~Sideris and L.~A. Rodriguez, ``{A Riccati approach for constrained linear
  quadratic optimal control},'' \emph{International Journal of Control},
  vol.~84, no.~2, pp. 370--380, 2 2011.

\bibitem{Camacho2007ModelControl}
E.~F. Camacho and C.~Bordons, \emph{{Model predictive control}}.\hskip 1em plus
  0.5em minus 0.4em\relax Springer, 2007.

\bibitem{Valero-Gomez2013}
A.~Valero-Gomez, J.~V. Gomez, S.~Garrido, and L.~Moreno, ``{The path to
  efficiency: Fast marching method for safer, more efficient mobile robot
  trajectories},'' \emph{IEEE Robotics and Automation Magazine}, vol.~20,
  no.~4, pp. 111--120, 12 2013.

\bibitem{Liu2015}
Y.~Liu and R.~Bucknall, ``{Path planning algorithm for unmanned surface vehicle
  formations in a practical maritime environment},'' \emph{Ocean Engineering},
  vol.~97, pp. 126--144, 3 2015.

\bibitem{Sanchez-Ibanez2019}
J.~R. S{\'{a}}nchez-Ib{\'{a}}nez, C.~J. P{\'{e}}rez-del Pulgar, M.~Azkarate,
  L.~Gerdes, and A.~Garc{\'{i}}a-Cerezo, ``{Dynamic path planning for
  reconfigurable rovers using a multi-layered grid},'' \emph{Engineering
  Applications of Artificial Intelligence}, vol.~86, pp. 32--42, 11 2019.

\bibitem{Bajracharya2008AutonomyFuture}
M.~Bajracharya, M.~W. Maimone, and D.~Helmick, ``{Autonomy for Mars Rovers:
  Past, present, and future},'' \emph{Computer}, vol.~41, no.~12, pp. 44--50,
  2008.

\bibitem{Merlo2013SampleMSR}
A.~Merlo, J.~Larranaga, and P.~Falkner, ``{Sample Fetching Rover (SFR) for
  MSR},'' in \emph{ASTRA 2013 - 12h Symposium on Advanced Space Technologies
  for Robotics and Automation}.\hskip 1em plus 0.5em minus 0.4em\relax European
  Space Agency (ESA), 2013.

\bibitem{Muirhead2019MarsConcepts}
B.~K. Muirhead and A.~Karp, ``{Mars Sample Return Lander Mission Concepts},''
  in \emph{IEEE Aerospace Conference Proceedings}, vol. 2019-March.\hskip 1em
  plus 0.5em minus 0.4em\relax IEEE Computer Society, 3 2019.

\bibitem{Azkarate2022DesignExploration}
M.~Azkarate, L.~Gerdes, T.~Wiese, M.~Zwick, M.~Pagnamenta, J.~Hidalgo~Carrio,
  P.~Poulakis, and C.~Perez-del Pulgar, ``{Design, Testing, and Evolution of
  Mars Rover Testbeds: European Space Agency Planetary Exploration},''
  \emph{IEEE Robotics {\&} Automation Magazine}, pp. 2--15, 1 2022.

\bibitem{Castilla-Arquillo2022Hardware-acceleratedSimulations}
R.~Castilla-Arquillo, C.~J. P{\'e}rez-del Pulgar, G.~J. Paz-Delgado, and
  L.~Gerdes, ``Hardware-accelerated mars sample localization via deep transfer
  learning from photorealistic simulations,'' \emph{arXiv preprint
  arXiv:2206.02622}, 2022.

\bibitem{Geiger2011StereoScan:Real-time}
A.~Geiger, J.~Ziegler, and C.~Stiller, ``{StereoScan: Dense 3d reconstruction
  in real-time},'' \emph{IEEE Intelligent Vehicles Symposium, Proceedings}, pp.
  963--968, 2011.

\bibitem{Mantoani2022SamplesRover}
L.~M. Mantoani, R.~Castilla-Arquillo, G.~J. Paz-Delgado, C.~J.
  P{\'{e}}rez-Del-Pulgar, and M.~Azkarate, ``{Samples Detection and Retrieval
  for a Sample Fetch Rover},'' in \emph{16th Symposium on Advanced Space
  Technologies in Robotics and Automation (ASTRA)}, Noordwijk, The Netherlands,
  2022.

\end{thebibliography}



\begin{IEEEbiography}[{\includegraphics[width=1in,height=1.25in,clip,keepaspectratio]{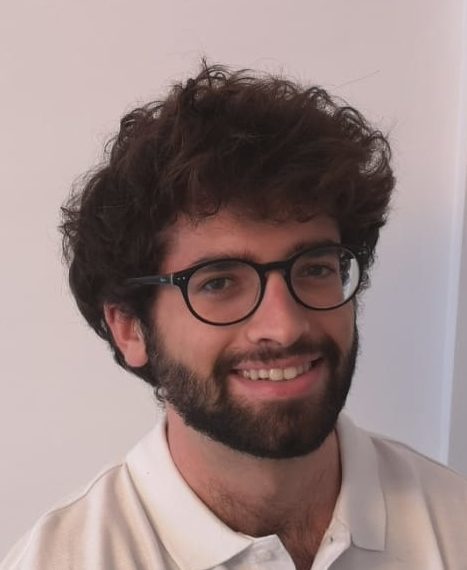}}]{Gonzalo J. Paz-Delgado} received his B.Sc. in industrial engineering, his M.Sc. in industrial engineering and his M.Sc. in mechatronics from the University of Malaga, Malaga, Spain, in 2018, 2021 and 2021 respectively. In 2018 he was granted an internship at the Space Robotics Laboratory, University of Malaga, including a research stay in the Planetary Robotics Laboratory, European Space Agency. Since 2019 he has a position as Young Researcher in the University of Malaga, participating in several projects related to autonomous navigation in planetary exploration vehicles. Besides, he is currently working on a co-supervised Ph.D. on mechatronics engineering, alongside DFKI and Universität Bremen, focused in motion planning for mobile manipulators. His research interests include robotics, motion planning, planetary exploration, automation and artificial intelligence.
\end{IEEEbiography}

\begin{IEEEbiography}[{\includegraphics[width=1in,height=1.25in,clip,keepaspectratio]{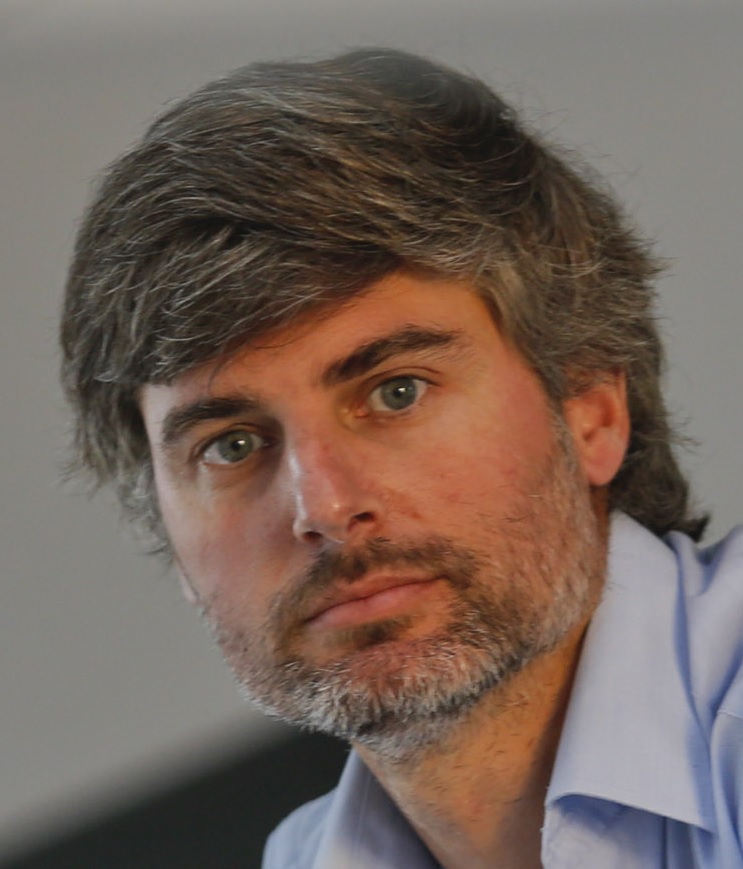}}]{Carlos P\'erez-Del-Pulgar}
received M.S. degree and PhD in Computer Science at the University of Málaga (UMA). In 2004, he was given a permanent position as research support staff at the University of Malaga until 2017. Also, since 2010 he was part-time assistant lecturer in the Electrical Engineering Faculty where he was responsible of different subjects related to automation and robotics. Currently he is associate lecturer and responsible of the Space Robotics Laboratory at the same University. His research interests include machine learning, surgical robotics, space robotics and autonomous astronomy. He has more than 60 publications in these topics and has been involved in more than 15 different Spanish and international projects. 
\end{IEEEbiography}

\begin{IEEEbiography}[{\includegraphics[width=1in,height=1.25in,clip,keepaspectratio]{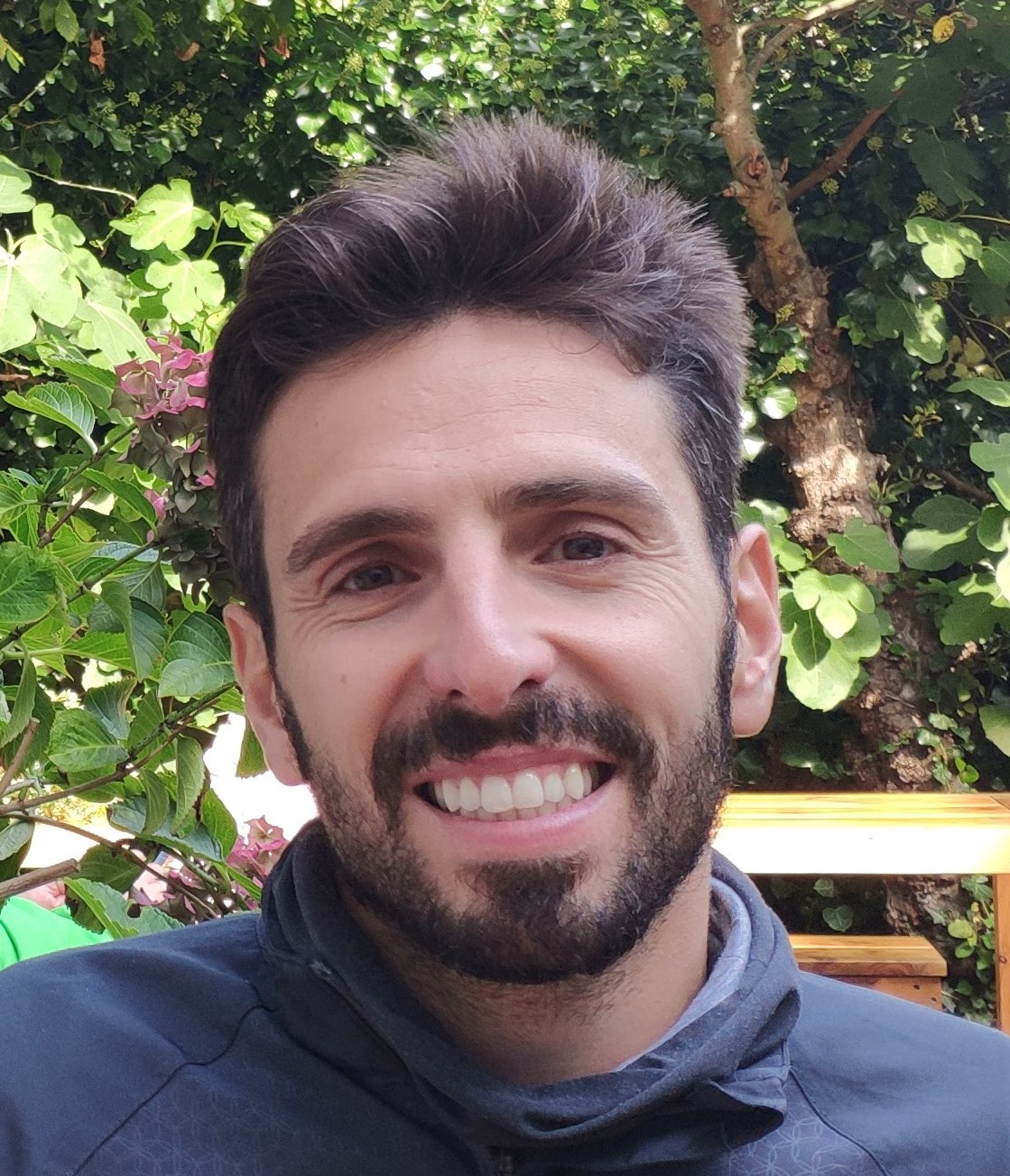}}]{Martin Azkarate} received his double M.S. degree in Electrical and Aerospace Engineering from Telecom BCN (UPC) and ISAE-Supaero respectively, after completing a research internship to conduct his Master Thesis at the Space Systems Lab of MIT. In 2012 he joined the Automation \& Robotics Section of European Space Agency (ESA), first as a trainee and with a permanent position as from 2014, where he has been working since in the area of robotics for planetary exploration, with particular focus on the autonomous navigation aspects of Mars rovers and being responsible of the Planetary Robotics Lab from 2014. In addition, Martin graduated with a PhD in Mechatronics at the University of Malaga (UMA) in December 2021, with the thesis entitled "Autonomous Navigation of Planetary Rovers". He has several papers and presentations in international conferences and high impact journals on works related to navigation, perception, planning and autonomy for rovers. In the last years he has participated as Technical Officer in several ESA and EU projects. Finally, from 2020, he has taken the role of GNC System Lead for the Sample Fetch Rover mission of the Mars Sample Return program.
\end{IEEEbiography}

\begin{IEEEbiography}[{\includegraphics[width=1in,height=1.25in,clip,keepaspectratio]{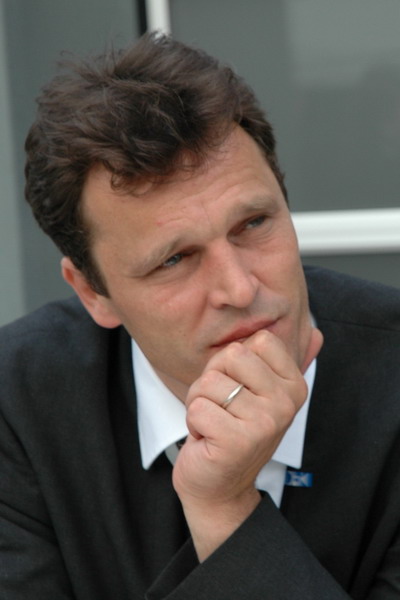}}]{Frank Kirchner} studied computer science and neurobiology at the University Bonn from 1989 until 1994, where he received his Dr. rer. nat. degree in computer science in 1999. Since 1994 he was Senior Scientist at the Gesellschaft für Mathematik und Datenverarbeitung (GMD) in Sankt Augustin and also a Senior Scientist at the Department for Electrical Engineering from 1998 at Northeastern University in Boston, MA, USA. In 1999, Dr. Kirchner first was appointed adjunct and then Tenure Track Assistant Professor at the Northeastern University, and since 2002 as a Full Professor at the University of Bremen. Since December 2005, Dr. Kirchner is also Director of the Robotics Innovation Center (RIC) in Bremen
\end{IEEEbiography}

\begin{IEEEbiography}[{\includegraphics[width=1in,height=1.25in,clip,keepaspectratio]{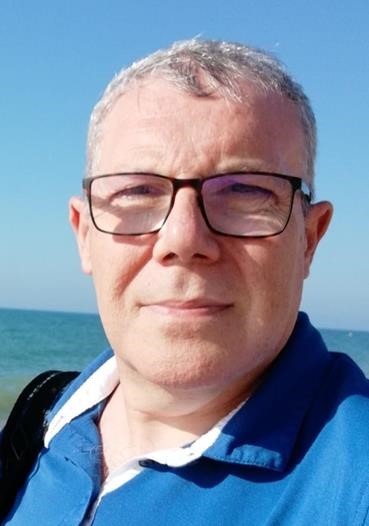}}]{Alfonso Garc\'ia-Cerezo} was born in 1959 in Sigüenza (Spain). He received the Industrial Electrical Engineer and the Doctoral Engineer Degree from the Escuela Tecnica Superior de Ingenieros Industriales of Vigo in 1983 and 1987, respectively.
From 1983 to 1988 he was Associate Professor in the Department of Electrical Engineering, Computers and Systems at the University of Santiago de Compostela. From 1988 to 1991 he was Assistant Professor at the same university.
Since 1992 he has been a Professor of System Engineering and Automation at the University of Malaga, Spain. Since 1993 to 2004 he has been Director of the “Escuela Tecnica Superior de Ingenieros Industriales de Malaga”. He is now the Head of the Department, and the responsible of the “Instituto de Automática Avanzada y Robótica de Andalucía” in Málaga.
He has authored or co-authored about 150 journal articles, conference papers, book chapters and technical reports. His current research included mobile robots and autonomous vehicles, surgical robotics, mechatronics and intelligent control.
Besides, is the responsible of more than 19 research projects during the last 10 years.
He is also a member of several international and national scientific and technical societies like IEEE, EUROBOTICS, IFAC,  AER-ATP, CEA and SEIDROB.
\end{IEEEbiography}

\end{document}